\documentclass[10pt,twocolumn,letterpaper]{article}
\usepackage[pagenumbers]{cvpr}
\usepackage{array}
\usepackage{multirow}
\usepackage{tabularx}
\usepackage{colortbl}
\usepackage{tcolorbox}
\tcbuselibrary{breakable,listings}
\definecolor{prompttitle}{gray}{0.23}
\definecolor{promptbody}{gray}{0.96}
\newtcblisting{promptbox}[1]{%
  listing only,breakable,sharp corners,lines before break=6,
  colback=promptbody,colframe=prompttitle,
  colbacktitle=prompttitle,coltitle=white,
  title={#1},fonttitle=\normalfont\normalsize\bfseries,
  boxrule=0.4pt,left=3pt,right=3pt,top=4pt,bottom=4pt,
  before skip=8pt,after skip=8pt,pad at break*=2pt,
  listing options={basicstyle=\normalfont\small\ttfamily,
    columns=fullflexible,keepspaces=true,breaklines=true,
    breakatwhitespace=true,breakindent=0pt,
    literate={_}{{\_\allowbreak}}1,
    showstringspaces=false,aboveskip=0pt,belowskip=0pt}
}
\makeatletter
\renewcommand{\paragraph}{%
  \@startsection{paragraph}{4}%
  {\z@}{-0.5em}{-0.5em}%
  {\normalfont\normalsize\bfseries}%
}
\makeatother
\AddToHook{env/tabularx/begin}{\footnotesize}
\AddToHook{env/tabular/begin}{\footnotesize}
\newcommand{\method}{EmbodiedBrain\xspace}
\newcommand{\ours}{\textit{(ours)}}
\definecolor{rankone}{gray}{0.85}
\definecolor{ranktwo}{gray}{0.93}
\definecolor{errorgrounding}{HTML}{00B0F0}
\definecolor{errorexecution}{HTML}{FA786B}
\definecolor{errorspatial}{HTML}{FFD039}
\definecolor{errorplanning}{HTML}{92D050}
\definecolor{errorenvironment}{HTML}{B192EF}
\newcommand{\best}[1]{\textbf{#1}}
\newcommand{\second}[1]{\underline{#1}}
\newcommand{\unreported}{\textemdash}
\newcolumntype{L}[1]{>{\raggedright\arraybackslash}p{#1}}

\definecolor{cvprblue}{rgb}{0.21,0.49,0.74}
\usepackage[pagebackref,breaklinks,colorlinks,allcolors=cvprblue]{hyperref}
\def\confName{CVPR}
\def\confYear{2027}
\title{EmbodiedBrain: 3D Visual Harness for Embodied Agents}
\author{%
Xinze~Du\textsuperscript{1,*},\quad Haokai~Zhang\textsuperscript{1,*,\textdagger},\quad Chi~Zhang\textsuperscript{1},\quad Yunze~Zhang\textsuperscript{1},\quad Zichong~Shi\textsuperscript{1},\quad Linhao~Zhong\textsuperscript{1}\\
Kaijun~Wang\textsuperscript{1},\quad Changtao~Miao\textsuperscript{3},\quad Chenchen~Jing\textsuperscript{4},\quad Yuhang~Ding\textsuperscript{1},\quad Liyang~Li\textsuperscript{1},\quad Muzhi~Zhu\textsuperscript{1}\\
Zhiyue~Zhao\textsuperscript{1,2},\quad Zongze~Du\textsuperscript{1},\quad Hao~Zhong\textsuperscript{1},\quad Yuling~Xi\textsuperscript{1,\textdaggerdbl},\quad Hao~Chen\textsuperscript{1,\textdaggerdbl},\quad Chunhua~Shen\textsuperscript{1,\textdaggerdbl}\\[5pt]
{\large\textsuperscript{1}Zhejiang University\enspace\textsuperscript{2}Shanghai Innovation Institute\enspace\textsuperscript{3}Ant Group\enspace\textsuperscript{4}Zhejiang University of Technology}\\[3pt]
{\small\textsuperscript{*}Equal contribution\qquad
\textsuperscript{\textdagger}Project leader\qquad
\textsuperscript{\textdaggerdbl}Corresponding authors}\\[3pt]
{\small Project Page: \url{https://aim-uofa.github.io/EmbodiedBrain/}}%
}
\hypersetup{pdfauthor={Xinze Du, Haokai Zhang, Chi Zhang, Yunze Zhang, Zichong Shi, Linhao Zhong, Kaijun Wang, Changtao Miao, Chenchen Jing, Yuhang Ding, Liyang Li, Muzhi Zhu, Zhiyue Zhao, Zongze Du, Hao Zhong, Yuling Xi, Hao Chen, Chunhua Shen}}
\newcommand{\teasercaption}{We introduce \textbf{EmbodiedBrain}, a 3D visual harness designed to unleash capabilities of Vision--Language Models (VLMs).}
\begin{document}
\onecolumn
\raggedbottom
\begingroup
\setlength{\topskip}{0pt}
\AddToHookNext{env/tabular/begin}{\large\setlength{\tabcolsep}{1pt}}
\makeatletter
\patchcmd{\@maketitle}{\vskip .5em \vspace*{12pt}}{}{}{}
\makeatother
\maketitle
{\centering
  \includegraphics[width=\textwidth,height=444bp,keepaspectratio]{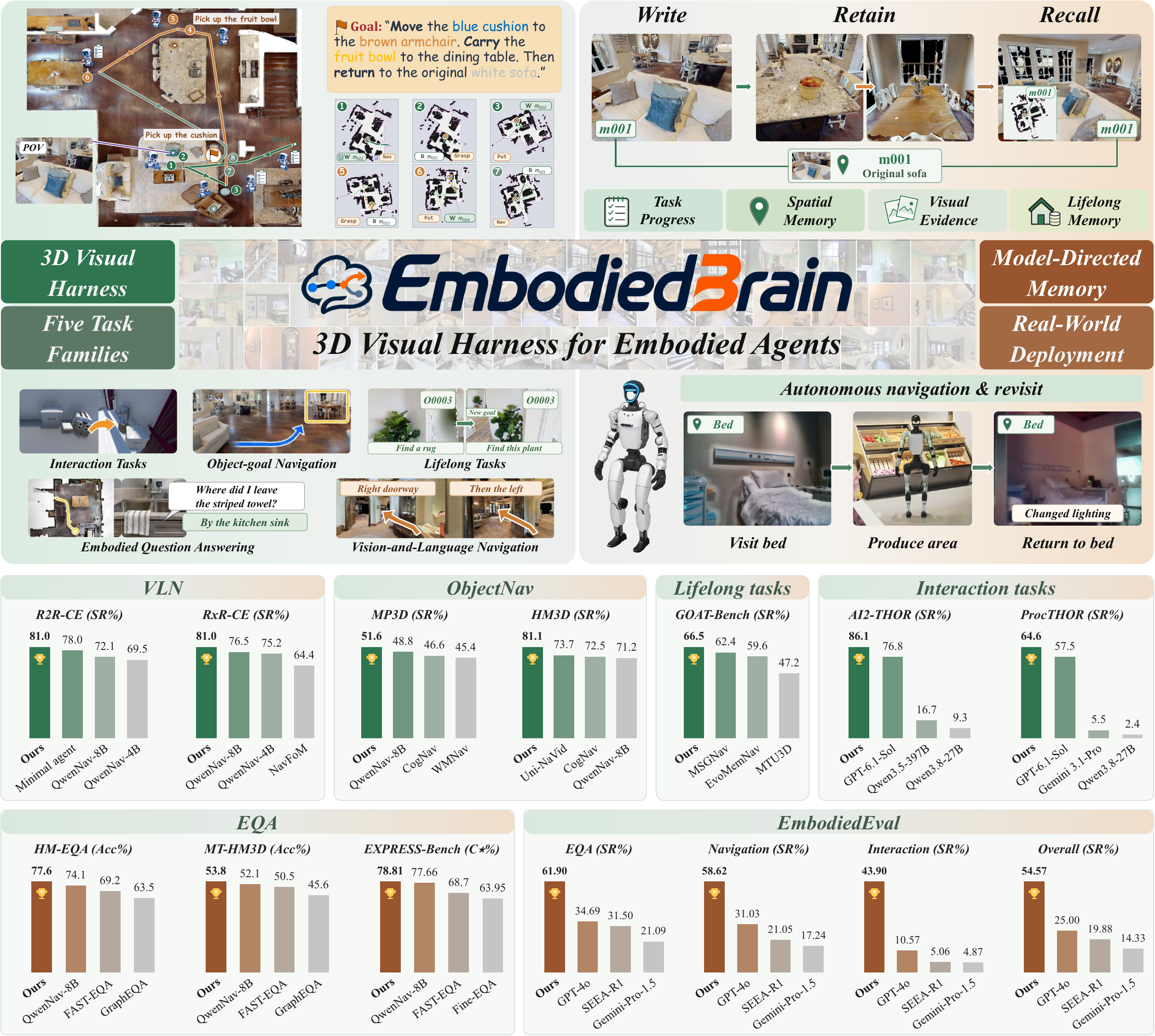}\par
  \captionsetup{hypcap=false}
  \captionof{figure}{\teasercaption}
  \label{fig:teaser}
}
\endgroup
\twocolumn
\flushbottom
\begin{abstract}
What do embodied agents call for in a 3D visual harness? \\To unleash the capabilities of existing vision--language models (VLMs) for embodied decision-making, we present \method, a 3D visual harness that combines a model-managed memory bank, a flexible joint action space and pluggable task interfaces.
The memory bank retains task-relevant information, including progress records, object locations, visual evidence and scene knowledge, while stored locations are visualized on bird's-eye views (BEV) alongside scene geometry and interaction history. Through joint action space, the model actively manages the memory bank and selects navigation targets by pointing in overhead view or BEV. Moreover, pluggable task interfaces then translate model actions into task-specific operations, adapting \method to multiple embodied tasks. 
Experiments across vision-and-language navigation, object-goal navigation, lifelong tasks, embodied question answering and interaction tasks show competitive results without task-specific policy fine-tuning. Deployment on a Unitree G1 further validates \method's effectiveness on a real-world long-horizon task requiring planning and memory. Further analyses offer insights into the design principles underlying its effectiveness. Code will be released.
\end{abstract}

\section{Introduction}
\label{sec:introduction}

\suppressfloats[t]
\begin{figure}[t]
\centering
\includegraphics[width=\columnwidth]{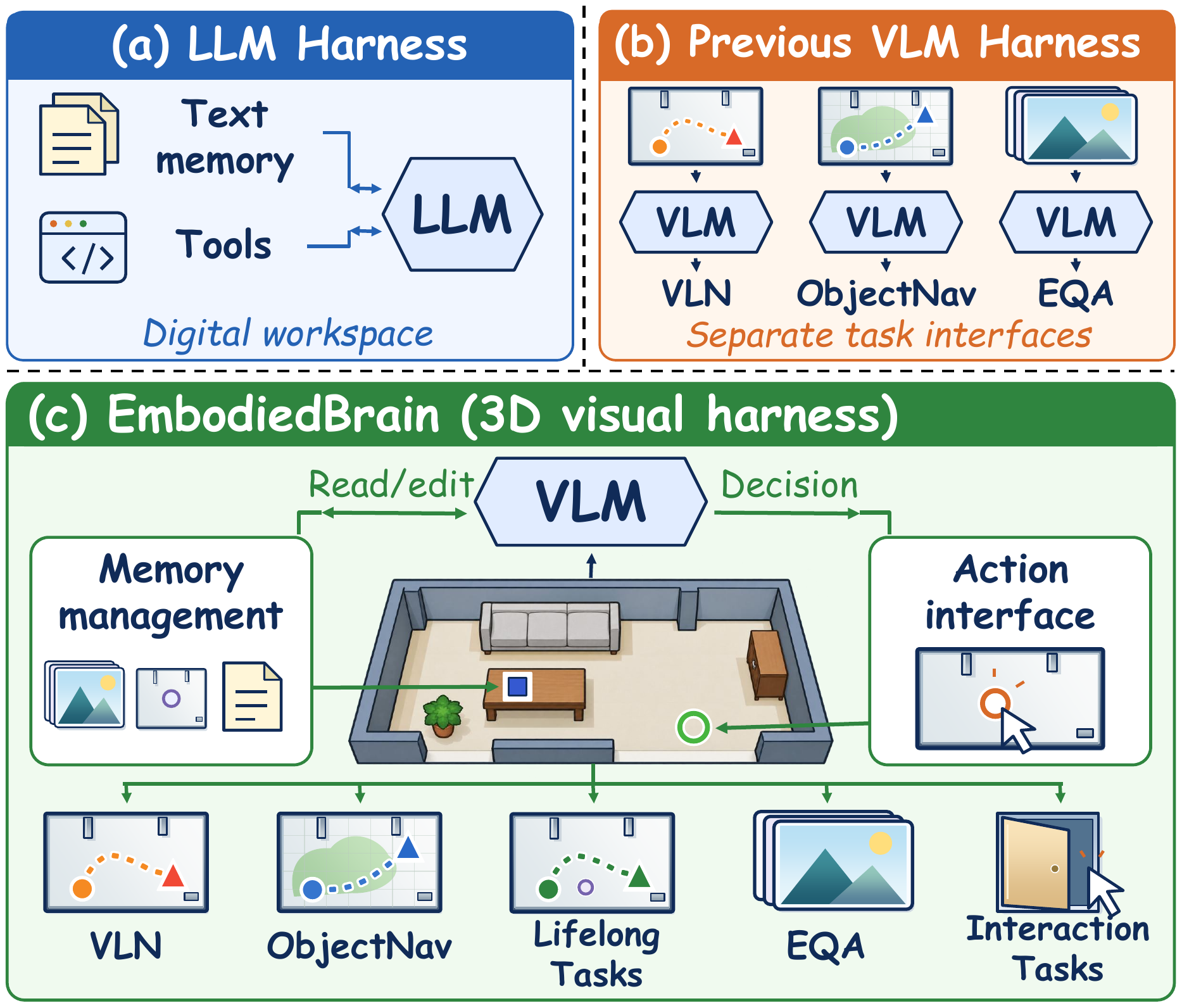}
\caption{Language-agent harnesses, task-focused embodied designs, and the shared 3D visual harness of EmbodiedBrain.}
\label{fig:harness_comparison}
\end{figure}

Embodied agents promise broad applications in household assistance~\cite{home_robot,dynamem,owmm}, warehouse automation~\cite{package_manipulation,warehouse_robotics2026,logistics_binpicking2025}, and autonomous inspection~\cite{anymal_field,autoinspect2025,qrivas2026}. Many of these applications involve long-horizon decision-making under partial observability, as agents move between locations, interact with the environment, and pursue successive subgoals. For LLM agents, text-based harnesses already support sustained interaction with their environment by organizing context, managing memory, and coordinating tool use~\cite{react,sweagent,mem0,memp,memharness,tool_orchestration,context_engineering}. Extending this role to embodied agents calls for a \textbf{3D visual harness} that enables Vision--Language Models (VLMs) to interact with 3D environments. We therefore ask: \textit{what do long-horizon embodied agents call for in a 3D visual harness?}

The design of a 3D visual harness involves three connected choices: (1) \textit{what should the agent see?} (2) \textit{what should the agent remember?} (3) \textit{how should the agent act?} Existing methods fall short of different parts of this design space. 
For observation space, most existing methods rely on current and historical egocentric images, without directly presenting a global map view to the VLM~\cite{navid,goal2pixel,agentvln}.
For action space, some approaches provide limited discrete action space such as candidate selection and local commands, constraining the model to directly specify navigation destinations~\cite{mapgpt,navid,mapnav}. Others predict targets by directly pointing in an egocentric point of view (POV) image, but leave memory management to predefined heuristic rules~\cite{goal2pixel,p2dnav,agentvln}. But frameworks with model-directed memory management remain tailored to particular tasks, lacking transfer capabilities~\cite{harnessvln,evolife}. Taken together, these limitations motivate an interface that gives the model richer control. Our guiding principle therefore is to \textit{unleash existing VLM capabilities as much as possible}, such as agentic capabilities~\cite{gpt6astra,gemini4argon,glm53flash,qwen38}, spatial reasoning~\cite{spatialvlm,spatialrgpt,metricbench} and visual grounding~\cite{robobrain25,molmo,molmo2,groundingpi}, through designed harness interfaces that make these strengths directly available for long-horizon embodied decision-making.

We introduce \textbf{\method}, a \emph{3D visual harness} that combines a model-managed memory bank, a flexible joint action space, and pluggable task interfaces within a shared, persistent workspace. The memory bank retains task-progress records, task-relevant spatial information and visual evidence, while lifelong scene memory preserves scene knowledge across goals. Spatial memory is displayed directly on Bird's-Eye View (BEV), alongside scene geometry and history trajectory. Through the joint action space, the agent manages memory bank by deciding what visual evidence to retain, which places to remember, and how to revise its task progress. Moreover, the agent can flexibly select navigation destinations by directly pointing on BEV or overhead (local top-down rendering of the reconstructed scene) views. Finally, pluggable task interfaces connect agent actions to task-specific operations, enabling \method to support diverse embodied tasks. Together, these interfaces let the model \textit{act on what it has remembered}. 

We evaluate \method across five task families: vision-and-language navigation~\cite{r2r,rxr,vlnce}, object-goal navigation~\cite{mp3d,hm3d}, lifelong tasks~\cite{goatbench}, embodied question answering~\cite{exploreeqa,memoryeqa,fine_eqa}, and interaction tasks~\cite{spatialworld,embodiedeval,procthor,ai2thor}. \method achieves strong results on multiple benchmarks without task-specific policy fine-tuning (\cref{fig:teaser}).  
Deployment on a Unitree G1 humanoid further demonstrates the effectiveness in real-world long-horizon execution requiring planning and memory reuse.
Additional analyses provide empirical guidance for designing 3D visual harnesses that support long-horizon embodied decision-making.
To sum up, our contributions are:
\begin{itemize}
  \item We introduce \method, a 3D visual harness for embodied agents that lets VLMs actively manage memory and select executable actions through a shared 3D visual interface across tasks.
  \item We evaluate \method across five embodied task families and obtain competitive results on multiple benchmarks without task-specific policy fine-tuning.
  \item We deploy EmbodiedBrain on a Unitree G1 humanoid and demonstrate long-horizon task execution that reuses spatial memory across successive goals.
  \item We provide detailed analysis of 3D visual harness design and offer insights into the design principles underlying \method's performance.
\end{itemize}

\section{Related Work}
\label{sec:related_work}

\paragraph{Embodied Agents.}
Embodied agents built on language and vision--language models span robotic manipulation and long-horizon planning~\cite{rpg,tango}. Manipulation-oriented agents develop reusable programs, skills, or action policies, but organize their decisions around robot execution interfaces~\cite{rats,era,shaper,agentic_robot,li2026roboharness}. The second line uses spatial memory and planning for exploration, navigation, and embodied question answering, yet primarily formulates actions around movement and evidence gathering~\cite{exploreeqa,3dmem,grapheqa,memoryeqa}. We follow this second line with a 3D visual harness, \method, designed to share memory and action interfaces across diverse embodied tasks.

\begin{figure*}[t]
\centering
\includegraphics[width=\textwidth]{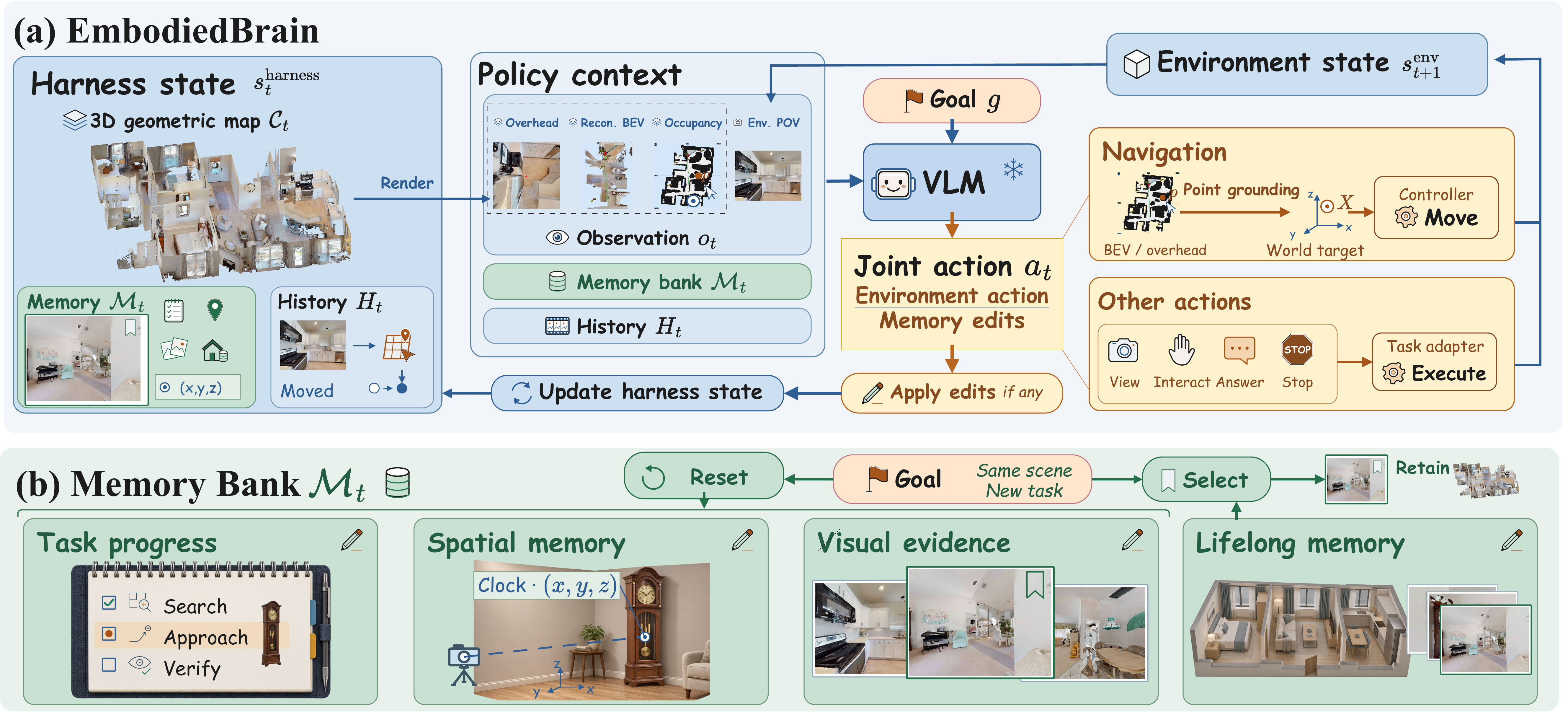}
\caption{\textbf{The 3D visual harness.} (a) A frozen VLM uses the goal, observations, memory, and history to select a joint action. Memory edits precede environment execution; measurements and feedback update the harness for the next decision. (b) The memory bank supports task progress, spatial knowledge, visual evidence, and lifelong scene reuse. For a new task in the same scene, task-local memories reset while the geometric map and scene memory persist.}
\label{fig:overview}
\end{figure*}

\paragraph{Agent Harnesses.}
LLM harnesses expose memory and tools to agents through text interfaces~\cite{react,memgpt,sweagent}. For embodied agents, however, the design of 3D visual harness remains fragmented, with limited systematic empirical guidance on how observations, memory, and actions should be organized.
For observation space, some approaches rely on current and historical egocentric images~\cite{navid,goal2pixel,agentvln}, while others provide spatial context through semantic maps or scene memories~\cite{mapnav,3dmem,vlmaps}. But these richer representations are often maintained through automatic aggregation rather than explicit model-directed edits~\cite{3dmem,mtu3d,vlmaps}.
For action space, some limit it to discrete commands~\cite{mapgpt,navid,mapnav}, while others predict egocentric point of view (POV) space action targets, but select historical images through visibility rules, fixed windows, or automatic sampling~\cite{goal2pixel,p2dnav,agentvln}.
Methods with adaptive retention or semantic writeback still focus on specific task families, with transferring capability beyond those families remaining limited~\cite{embodiednavigator,navharness_adaptive}. 
EmbodiedBrain provides a shared 3D visual interface through which VLMs actively manage memory and select executable actions across multiple embodied tasks.

\section{Methodology}
\label{sec:method}

\subsection{Preliminaries}
\label{sec:preliminaries}
Embodied agents interact with environments through actions and partial visual observations from environment state. We formulate this interaction as a partially observable Markov decision process (POMDP)~\cite{astrom1965optimal,kaelbling1998planning}, $\mathcal{P}=(\mathcal{S},\mathcal{A},\mathcal{O},T,\Omega)$. $\mathcal{S}=\mathcal{S}^{\mathrm{env}}\times\mathcal{S}^{\mathrm{harness}}$ is the joint state space, where $\mathcal{S}^{\mathrm{env}}$ and $\mathcal{S}^{\mathrm{harness}}$ are the environment and harness state spaces. At step $t$, the joint state $s_t=(s_t^{\mathrm{env}},s_t^{\mathrm{harness}})\in\mathcal{S}$ combines $s_t^{\mathrm{env}}\in\mathcal{S}^{\mathrm{env}}$ with $s_t^{\mathrm{harness}}=(\mathcal{C}_t,\mathcal{M}_t,H_t)\in\mathcal{S}^{\mathrm{harness}}$, where $\mathcal{C}_t$ is the 3D geometric map incrementally updated from posed RGB-D measurements, $\mathcal{M}_t$ is the memory bank managed by the frozen VLM policy $\pi_\theta$, and $H_t$ retains recent actions, observation and execution outcome from the last step. Following prior works~\cite{peshkin1999external,allen2024lambda}, the joint action space $\mathcal{A}=\mathcal{A}^{\mathrm{env}}\times\mathcal{A}^{\mathrm{mem}}$ combines environment interaction action $a_t^{\mathrm{env}}\in\mathcal{A}^{\mathrm{env}}$ and memory edit action $a_t^{\mathrm{mem}}\in\mathcal{A}^{\mathrm{mem}}$. The agent's visual observation space is $\mathcal{O}$, with observation kernel $\Omega$.

At step $t$, the agent receives a visual observation $o_t\in\mathcal{O}$ sampled from $\Omega(\,\cdot\mid s_t)$. Given a goal $g$, $\pi_\theta$ uses $o_t$ and $\mathcal{M}_t,H_t$ to produce $a_t\in\mathcal{A}$:
\begin{equation}
a_t=(a_t^{\mathrm{env}},a_t^{\mathrm{mem}})\sim\pi_\theta(\,\cdot\mid\mathcal{M}_t,H_t,o_t;g).
\label{eq:decision}
\end{equation}
The transition kernel $T$ produces $s_{t+1}\sim T(\,\cdot\mid s_t,a_t)$, updating both $s_{t+1}^{\mathrm{env}}$ and $s_{t+1}^{\mathrm{harness}}=(\mathcal{C}_{t+1},\mathcal{M}_{t+1},H_{t+1})$. The next observation is $o_{t+1}\sim\Omega(\,\cdot\mid s_{t+1})$, closing the loop.

\subsection{Observation Interface}
\label{sec:spatial_context}
To answer \textit{what the agent should see} under partial observability, the observation kernel $\Omega$ exposes both local egocentric detail and global 3D layout in a shared visual workspace, avoiding forcing the VLM to mentally reconstruct scene geometry from POV frames alone. Specifically, the environment provides the current POV image $I_{\text{pov},t}$, while the overhead image $I_{\text{oh},t}$ (local top-down renderings of the reconstructed scene), occupancy BEV $I_{\text{bev},t}$, and RGB reconstruction BEV $I_{\text{rec},t}$ are rendered from $\mathcal{C}_t$.

Furthermore, we overlay policy annotations $r_i$ and grounded evidence $e_j$ on the BEV, together with the past trajectory, the camera position at the previous step, and exploration candidates. 
Anchoring memory annotations and interaction history directly onto these aligned views lets the VLM exploit cross-view spatial correspondence when choosing destinations.
Thus the resulting visual observation is $o_t=\bigl\{I_{\mathrm{pov},t},I_{\mathrm{oh},t},I_{\mathrm{bev},t},I_{\mathrm{rec},t}\bigr\}\in\mathcal{O}$.

\subsection{Memory Bank}
\label{sec:memory}

To address \textit{what the agent should remember} over long horizon and successive goals, memory bank $\mathcal{M}_t = (\mathcal{M}^{\text{prog}}_t, \mathcal{M}^{\text{sp}}_t, \mathcal{M}^{\text{ve}}_t, \mathcal{M}^{\text{scene}}_t)$ is designed as a model-managed, visually grounded store rather than a passive heuristic cache or geometry-free text log. It retains complementary task context and spatial experience within an episode while preserving scene knowledge across tasks.

\paragraph{Task-progress memory.}
$\mathcal{M}_t^{\mathrm{prog}}$ organizes remaining task requirements through the current plan, progress assessments, task-relevant findings, and missing evidence, providing the task context for interpreting other memory components. The policy $\pi_\theta$ revises $\mathcal{M}_t^{\mathrm{prog}}$ using observations and execution feedback while interacting with the environment.

\paragraph{Spatial memory.}
$\mathcal{M}_t^{\mathrm{sp}}=\{r_i\}_i\cup\{e_j\}_j$ complements task-level memory $\mathcal{M}_t^{\mathrm{prog}}$ with spatial references for task-relevant clues through policy annotations $r_i$ and grounded evidence $e_j$. 
\(\pi_\theta\) selects an object or location by pointing in POV, BEV, or overhead image with a description \(d_i\), which is then mapped to a world location \(X_i\in\mathbb{R}^3\) using source-view geometry, storing $r_i=(X_i,d_i)$.
An external grounding model proposes possibly relevant objects from sampled observations, which is then mapped to a world coordinate in the same way with $r_i$, and each entry is represented as $e_j=(\rho_j,X_j,d_j)$, where $\rho_j\in\mathrm{SE}(3)$ is the observing camera pose, $X_j\in\mathbb{R}^3$ is the estimated object location, and $d_j$ is its semantic description. These references guide subsequent revisits and further inspection.

\paragraph{Visual evidence memory.}
$\mathcal{M}_t^{\mathrm{ve}}=\{(I_i,\rho_i,d_i)\}_i$ retains images for inspecting visual details beyond concise spatial memory $\mathcal{M}_t^{\mathrm{sp}}$. Its $i$-th entry contains the stored image $I_i$, the source camera pose $\rho_i\in\mathrm{SE}(3)$, and description $d_i$. $\pi_\theta$ actively saves and deletes entries to preserve evidence for subsequent inspection. The capacity $B_{\mathrm{ve}}$ bounds the number of entries, $|\mathcal{M}_t^{\mathrm{ve}}|\leq B_{\mathrm{ve}}$.

\begin{figure*}[!t]
\centering
\includegraphics[
  width=\textwidth,
  pagebox=cropbox,
  clip
]{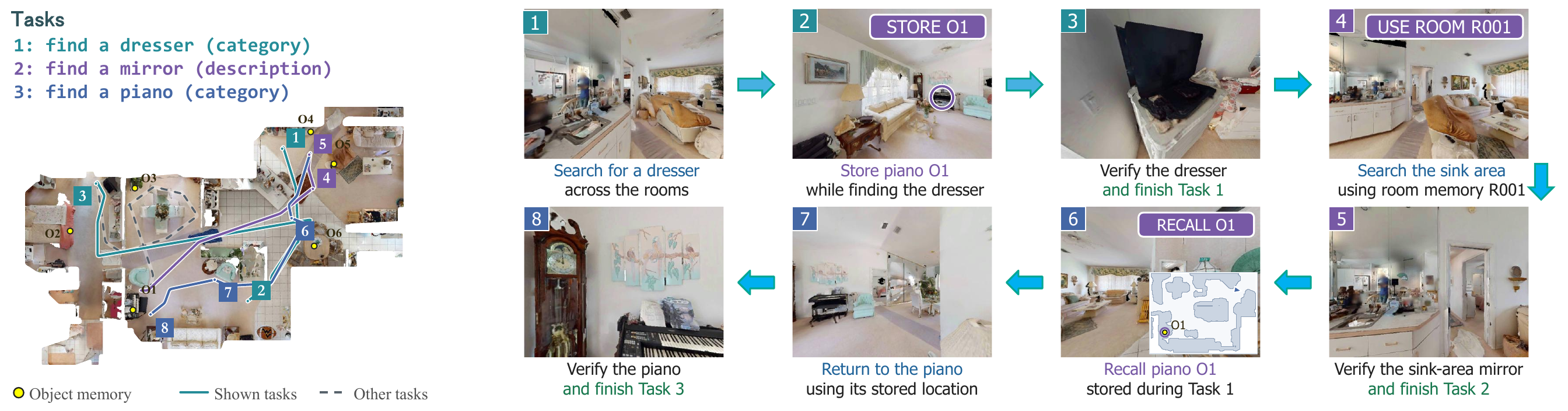}
\caption{\textbf{Lifelong tasks on GOAT-Bench.} While searching for a dresser, the agent stores the location of a piano. It then uses room memory to locate a described mirror and later recalls the piano's stored location to complete a new category goal.}
\label{fig:case_study}
\end{figure*}

\paragraph{Lifelong scene memory.}
\label{sec:lifelong_memory}
While $\mathcal{M}_t^{\mathrm{prog}}$, $\mathcal{M}_t^{\mathrm{sp}}$ and $\mathcal{M}_t^{\mathrm{ve}}$ organize information for the current task, $\mathcal{M}_t^{\mathrm{scene}}$ preserves grounded region and object knowledge, including locations, semantic descriptions, and associated visual evidence to reuse scene knowledge beyond the current task. $\mathcal{M}_t^{\mathrm{scene}}$ is continually updated as the agent explores within the same scene. When a new task $g$ begins in the same scene, the harness reinitializes $\mathcal{M}_t^{\mathrm{prog}}$, $\mathcal{M}_t^{\mathrm{sp}}$, and $\mathcal{M}_t^{\mathrm{ve}}$, and retains $\mathcal{C}_t$ and $\mathcal{M}_t^{\mathrm{scene}}$, while $\pi_\theta$ selects task-relevant information from $\mathcal{M}_t^{\mathrm{scene}}$ to guide decision-making.

\subsection{Action Interface}
\label{sec:actions}

Addressing \textit{how the agent should act}, the action interface couples environment execution with active memory management so that the VLM can directly act on what it sees and remembers.
Each environment action $a_t^{\mathrm{env}}=(\tau_t,\xi_t)\in\mathcal{A}^{\mathrm{env}}$ specifies an action type $\tau_t$ and its arguments $\xi_t$. Supported types include navigation, viewpoint changes, object interaction, answering, and termination, depending on the task adapter. Specifically, to make the spatial reasoning and pointing capabilities of VLM actionable, we let $\pi_\theta$ choose navigation destinations $p\in\mathbb{R}^2$ by pointing directly on the BEV/overhead images. For the selected source view $v$, the grounding map $\Gamma_v$ transforms $p$ into a world target $X\in\mathbb{R}^3$ as $X=\Gamma_v(p;s_t)$. The corresponding navigation controller checks traversability, plans a path to \(X\), and executes low-level environment actions decomposed from $a_t^{\mathrm{env}}$. 
Rather than constraining $\pi_\theta$ to myopic discrete commands, letting it point directly on BEV/overhead views converts VLM visual grounding into actionable 3D waypoints, while offloading low-level obstacle avoidance and path execution to the navigation controller.
Other actions such as interaction, answer, and completion actions are mapped to operations supported by each environment by task adapters.

To turn memory curation into an agentic decision rather than a fixed rule,
$\pi_\theta$ also actively manages the memory bank $\mathcal{M}_t$ through memory edit actions $a_t^{\mathrm{mem}}\in\mathcal{A}^{\mathrm{mem}}$, choosing how to revise task progress, which locations to annotate, and which visual evidence to save or delete.
The harness applies these edits before environment execution and updates \(\mathcal{C}_t\) and \(H_t\) from accepted RGB-D measurements and execution outcomes before the next decision.

\begin{table}[t]
\centering
\caption{\textbf{Vision-and-language navigation.} SR/SPL (\%). Training-based methods use full val-unseen splits; $\dagger$: zero-shot methods use the same 100-episode subset on R2R-CE~\cite{astra_vln,agents_control,harnessvln}. $\ddagger$: \method uses 100-episode subset on RxR-CE. ``\unreported'' denotes unreported metrics.}
\label{tab:vln}
\setlength{\tabcolsep}{3pt}
\begin{tabularx}{\linewidth}{>{\raggedright\arraybackslash}p{0.40\linewidth}*{4}{>{\centering\arraybackslash}X}}
\toprule
\multirow{2}{*}{\textbf{Method}} & \multicolumn{2}{c}{\textbf{R2R-CE}} & \multicolumn{2}{c}{\textbf{RxR-CE}}\\
\cmidrule(lr){2-3}\cmidrule(lr){4-5}
& SR $\uparrow$ & SPL $\uparrow$ & SR $\uparrow$ & SPL $\uparrow$\\
\midrule
\multicolumn{5}{l}{\textit{Training-based methods}}\\
Uni-NaVid~\cite{uninavid} & 47.0 & 42.7 & 48.7 & 40.9\\
NaVILA~\cite{navila} & 54.0 & 49.0 & 49.3 & 44.0\\
JanusVLN~\cite{janusvln} & 60.5 & 56.8 & 56.2 & 47.5\\
NavFoM~\cite{navfom} & 61.7 & 55.3 & 64.4 & 56.2\\
DualVLN~\cite{dualvln} & 64.3 & 58.5 & 61.4 & 51.8\\
OmniNav~\cite{omninav} & 69.5 & 66.1 & 73.6 & 62.0\\
QwenNav-4B~\cite{qwenrobotnav} & 69.5 & 63.6 & 75.2 & \second{65.0}\\
QwenNav-8B~\cite{qwenrobotnav} & 72.1 & \second{66.6} & \second{76.5} & \best{65.7}\\
\midrule
\multicolumn{5}{l}{\textit{Zero-shot methods}}\\
Open-Nav$^\dagger$~\cite{opennav} & 32.0 & 27.7 & \unreported & \unreported\\
EvoNav$^\dagger$~\cite{evonav} & 43.0 & 37.8 & \unreported & \unreported\\
SmartWay$^\dagger$~\cite{smartway} & 44.0 & 35.0 & \unreported & \unreported\\
AgenticNav$^\dagger$~\cite{agenticnav} & 55.0 & 48.4 & \unreported & \unreported\\
HarnessVLN$^\dagger$~\cite{harnessvln} & 58.0 & 35.8 & 51.4 & 38.1\\
Minimal agent$^\dagger$~\cite{agents_control} & \second{78.0} & 65.3 & \unreported & \unreported\\
\rowcolor{ranktwo}
\textbf{\method}$^{\dagger\ddagger}$ \ours & \best{81.0} & \best{70.2} & \best{81.0} & 62.8\\
\bottomrule
\end{tabularx}
\end{table}

\section{Experiments}
\label{sec:experiments}
We evaluate \method across five task families, and deploy it on a Unitree G1 to finish real-world tasks. Detailed benchmark and baseline introductions, and experimental setups are provided in the supplementary material.

\subsection{Vision-and-language navigation}
\label{sec:vln_results}
Following a route instruction requires the agent to relate successive landmarks and spatial constraints to its changing position. We evaluate this kind of ability of \method on R2R-CE and RxR-CE \cite{r2r,rxr,vlnce}.
Due to the complexity of VLN, we use GPT-6-Astra~\cite{gpt6astra} as our base VLM. To control inference costs, we evaluate R2R-CE on the ON-100 subset of Open-Nav~\cite{opennav}, following prior works~\cite{opennav,agenticnav,astra_vln,harnessvln}. For RxR, we randomly sample 100 English guide episodes from val-unseen.

\begin{table*}[t]
\centering
\caption{\textbf{Results on EmbodiedEval across task families.} SR and goal-condition success (GcS) are percentages. Specifically, to align with our evaluation task families, we group the original five task subsets of EmbodiedEval~\cite{embodiedeval} into three categories: navigation, EQA, and interaction. 
SEEA-R1~\cite{seea_r1} uses its ground-truth-reward variant. ``\unreported" denotes unreported metrics.}
\label{tab:embodiedeval}
\setlength{\tabcolsep}{3pt}
\begin{tabularx}{\textwidth}{@{}>{\raggedright\arraybackslash}p{0.24\textwidth}*{7}{>{\centering\arraybackslash}X}@{}}
\toprule
\multirow{2}{*}{\textbf{Method}} & \textbf{EQA} & \multicolumn{2}{c}{\textbf{Navigation}} & \multicolumn{2}{c}{\textbf{Interaction}} & \multicolumn{2}{c}{\textbf{Overall}}\\
\cmidrule(lr){2-2}\cmidrule(lr){3-4}\cmidrule(lr){5-6}\cmidrule(lr){7-8}
& SR $\uparrow$ & SR $\uparrow$ & GcS $\uparrow$ & SR $\uparrow$ & GcS $\uparrow$ & SR $\uparrow$ & GcS $\uparrow$\\
\midrule
GPT-4o & \second{34.69} & \second{31.03} & \second{42.53} & \second{10.57} & \second{24.93} & \second{25.00} & \second{32.42}\\
Gemini-Pro-1.5 & 21.09 & 17.24 & 25.86 & 4.87 & 13.96 & 14.33 & 19.26\\
SEEA-R1~\cite{seea_r1} & 31.50 & 21.05 & \unreported & 5.06 & \unreported & 19.88 & \unreported\\
\midrule
\rowcolor{ranktwo}
\textbf{\method} \ours & \best{61.90} & \best{58.62} & \best{65.52} & \best{43.90} & \best{57.72} & \best{54.57} & \best{61.13}\\
\bottomrule
\end{tabularx}
\end{table*}

On R2R-CE, \method leads the methods in \cref{tab:vln} in both SR and SPL, improving over the strongest zero-shot baseline, the Minimal agent~\cite{agents_control}, by 3.0 and 4.9 percentage points. For RxR-CE, our sampled tasks yield 81.0\% SR and 62.8\% SPL; HarnessVLN~\cite{harnessvln} reports 51.4\% and 38.1\% on full val-unseen. The training-based QwenNav-8B~\cite{qwenrobotnav} reports 76.5\% SR and 65.7\% SPL on the full set. Thus, \method has higher success on its samples.

\subsection{Object-goal navigation}
\label{sec:objectnav_results}
Object-goal navigation tests \method's ability to locate objects in unseen scenes on MP3D and HM3D-v2~\cite{mp3d,hm3d}, with an additional test on EmbodiedEval's navigation subset~\cite{embodiedeval}. We use Qwen3.8-27B as the base VLM. 
On MP3D, \method surpasses all the baselines in both metrics, with 51.6\% SR and 18.8\% SPL (\cref{tab:objectnav}). On HM3D, it achieves 81.1\% SR, 7.4 percentage points above the strongest listed baseline, Uni-NaVid~\cite{uninavid}. The comparison with GPT-4o with official naive harness on EmbodiedEval~\cite{embodiedeval} also favors \method in both metrics: it completes 34 of 58 navigation tasks, giving 58.62\% SR and 65.52\% GcS (Goal-Condition Success) (\cref{tab:embodiedeval}).

\begin{table}[t]
\centering
\caption{\textbf{Object-goal navigation.} SR/SPL (\%). $\dagger$: evaluate on HM3D-v1; other methods use harder HM3D-v2.
``\unreported" denotes unreported metrics.}
\label{tab:objectnav}
\setlength{\tabcolsep}{3pt}
\begin{tabularx}{\linewidth}{>{\raggedright\arraybackslash}p{0.40\linewidth}*{4}{>{\centering\arraybackslash}X}}
\toprule
\multirow{2}{*}{\textbf{Method}} & \multicolumn{2}{c}{\textbf{MP3D}} & \multicolumn{2}{c}{\textbf{HM3D}}\\
\cmidrule(lr){2-3}\cmidrule(lr){4-5}
& SR $\uparrow$ & SPL $\uparrow$ & SR $\uparrow$ & SPL $\uparrow$\\
\midrule
VLFM$^{\dagger}$~\cite{vlfm} & 36.4 & 17.5 & 52.5 & 30.4\\
OpenFMNav$^{\dagger}$~\cite{openfmnav} & 37.2 & 15.7 & 52.5 & 24.1\\
SG-Nav$^{\dagger}$~\cite{sgnav} & 40.2 & 16.0 & 54.0 & 24.9\\
TriHelper$^{\dagger}$~\cite{trihelper} & \unreported & \unreported & 56.5 & 25.3\\
WMNav$^{\dagger}$~\cite{wmnav} & 45.4 & 17.2 & 58.1 & 31.2\\
CogNav$^{\dagger}$~\cite{cognav} & 46.6 & 16.1 & 72.5 & 26.2\\
Uni-NaVid$^{\dagger}$~\cite{uninavid} & \unreported & \unreported & \second{73.7} & \best{37.1}\\
QwenNav-8B~\cite{qwenrobotnav} & \second{48.8} & \second{17.7} & 71.2 & 33.0\\
STEGNav~\cite{stegnav} & \unreported & \unreported & 69.4 & 28.2\\
\midrule
\rowcolor{ranktwo}
\textbf{\method} \ours & \best{51.6} & \best{18.8} & \best{81.1} & \second{34.2}\\
\bottomrule
\end{tabularx}
\end{table}

\subsection{Lifelong tasks}
\label{sec:lifelong_results}
We further examine whether \method can solve lifelong tasks by reusing scene knowledge across successive goals on the val-unseen split of GOAT-Bench~\cite{goatbench} with Qwen3.8-27B, retaining lifelong scene memory across category, description, and image goals within an episode. 
\method records 66.5\% SR and 46.1\% SPL (\cref{tab:goat}). These exceed the best listed baseline for each metric by 4.1 percentage points in SR over MSGNav~\cite{msgnav} and 7.2 in SPL over EvoMemNav~\cite{evomemnav}. 
\Cref{sec:what_should_agent_remember} examines the contribution of scene memory to lifelong tasks.
\Cref{fig:case_study} is an example that the agent reuses scene-memory to resolve several tasks within the same scene.

\subsection{Embodied question answering}
\label{sec:eqa_results}

Moreover, we assess \method's ability to gather visual evidence and answer questions on HM-EQA~\cite{exploreeqa}, MT-HM3D~\cite{memoryeqa}, EXPRESS-Bench~\cite{fine_eqa}, and EmbodiedEval's EQA subset~\cite{embodiedeval}. The base VLM is Qwen3.8-27B for EmbodiedEval and Qwen3.8-Flash-Next for the others. With 77.6\% accuracy on HM-EQA and 53.8\% on MT-HM3D, \method ranks above QwenNav-8B~\cite{qwenrobotnav} and other baselines in \cref{tab:eqa}. This advantage comes with higher normalized steps: 0.20 and 0.24, vs QwenNav-8B's 0.17 and 0.22. 
For EXPRESS-Bench, \method obtains 78.81 in answer quality and 40.27 in path efficiency. The advantage over QwenNav-8B is larger for path efficiency than answer quality: 8.54\% versus 1.15\%. EmbodiedEval further tests attribute and spatial questions and \method gets 61.90\% SR and surpasses all the baselines (\cref{tab:embodiedeval}).

\begin{table}[t]
\centering
\caption{\textbf{Lifelong tasks on GOAT-Bench.} SR/SPL (\%) on val-unseen. ``\unreported'' denotes unreported metrics.}
\label{tab:goat}
\begin{tabularx}{\linewidth}{>{\raggedright\arraybackslash}p{0.40\linewidth}*{2}{>{\centering\arraybackslash}X}}
\toprule
\multirow{2}{*}[-2pt]{\textbf{Method}} & \multicolumn{2}{c}{\textbf{GOAT-Bench}}\\
\cmidrule(lr){2-3}
& SR $\uparrow$ & SPL $\uparrow$\\
\midrule
Modular GOAT~\cite{goatbench} & 24.9 & 17.2\\
VLMnav~\cite{vlmnav} & 20.1 & 9.6\\
DyNaVLM~\cite{dynavlm} & 25.5 & 10.2\\
TANGO~\cite{tango} & 32.1 & 16.5\\
MTU3D~\cite{mtu3d} & 47.2 & 27.7\\
3D-Mem~\cite{3dmem} & 42.6 & 22.8\\
EvoMemNav~\cite{evomemnav} & 59.6 & \second{38.9}\\
MSGNav~\cite{msgnav} & \second{62.4} & \unreported\\
\midrule
\rowcolor{ranktwo}
\textbf{\method} \ours & \best{66.5} & \best{46.1}\\
\bottomrule
\end{tabularx}
\end{table}

\begin{table}[t]
\centering
\caption{\textbf{Embodied question answering.} Accuracy, answer score $C^{\star}$, and path efficiency $E_{\mathrm{path}}$ are percentages. Steps are reported normalized equivalent steps. 
$\dagger$: Qwen3.5-Plus planner with a QwenNav-8B executor. ``\unreported" denotes unreported metrics.}
\label{tab:eqa}
\setlength{\tabcolsep}{2pt}
\begin{tabularx}{\linewidth}{@{}>{\raggedright\arraybackslash}X*{6}{c}@{}}
\toprule
\multirow{2}{*}{\textbf{Method}} & \multicolumn{2}{c}{\mbox{\textbf{HM-EQA}}} & \multicolumn{2}{c}{\mbox{\textbf{MT-HM3D}}} & \multicolumn{2}{c}{\textbf{EXPRESS}}\\
\cmidrule(lr){2-3}\cmidrule(lr){4-5}\cmidrule(lr){6-7}
& Acc. $\uparrow$ & Steps $\downarrow$ & Acc. $\uparrow$ & Steps $\downarrow$ & $C^{\star}$ $\uparrow$ & $E_{\mathrm{path}}$ $\uparrow$\\
\midrule
Explore-EQA~\cite{exploreeqa} & 58.4 & 0.52 & 36.2 & 0.64 & \unreported & \unreported\\
GraphEQA~\cite{grapheqa} & 63.5 & \second{0.20} & 45.6 & 0.45 & \unreported & \unreported\\
MemoryEQA~\cite{memoryeqa} & 61.4 & 0.40 & 43.1 & 0.41 & \unreported & \unreported\\
Fine-EQA~\cite{fine_eqa} & 56.0 & 0.54 & \unreported & \unreported & 63.95 & 25.58\\
3D-Mem~\cite{3dmem} & 50.4 & 0.63 & \unreported & \unreported & \unreported & \unreported\\
FAST-EQA~\cite{fasteqa} & 69.2 & 0.65 & 50.5 & 0.52 & 68.70 & 29.25\\
QwenNav-8B$^{\dagger}$~\cite{qwenrobotnav} & \second{74.1} & \best{0.17} & \second{52.1} & \best{0.22} & \second{77.66} & \second{31.73}\\
\midrule
\rowcolor{ranktwo}
\textbf{\method} \ours & \best{77.6} & \second{0.20} & \best{53.8} & \second{0.24} & \best{78.81} & \best{40.27}\\
\bottomrule
\end{tabularx}
\end{table}

\subsection{Interaction tasks}
\label{sec:interaction_results}
Finally, we test \method's ability to fulfill instructions through object-state changes or interactions with people on AI2-THOR and ProcTHOR subsets of SpatialWorld~\cite{spatialworld,procthor,ai2thor} and EmbodiedEval's interaction subset~\cite{embodiedeval}. Using Qwen3.8-27B, changing from SpatialWorld's native harness to \method raises SR from 9.3\% to 86.1\% on AI2-THOR and from 2.4\% to 64.6\% on ProcTHOR (\cref{tab:spatialworld}). \method also outperforms GPT-6.1-Sol operating under the native harness on both subsets. EmbodiedEval's tasks extend the comparison to interactions with other agents as well as objects. \method also outperforms all the baselines (\cref{tab:embodiedeval}).

\begin{table}[t]
\centering
\caption{\textbf{Interaction tasks.} SR (\%) on AI2-THOR and ProcTHOR subsets of SpatialWorld.}
\label{tab:spatialworld}
\begin{tabularx}{\linewidth}{>{\raggedright\arraybackslash}p{0.40\linewidth}*{2}{>{\centering\arraybackslash}X}}
\toprule
\multirow{2}{*}[-2pt]{\textbf{Method}} & \textbf{AI2-THOR} & \textbf{ProcTHOR}\\
\cmidrule(lr){2-2}\cmidrule(lr){3-3}
& SR $\uparrow$ & SR $\uparrow$\\
\midrule
Gemini-3.1-Pro & 6.4 & 5.5\\
Qwen-3.5-397B-A17B & 16.7 & 0.0\\
Qwen3.8-27B & 9.3 & 2.4\\
GPT-5 & 16.1 & 0.0\\
GPT-6.1-Sol & \second{76.8} & \second{57.5} \\

\midrule
\rowcolor{ranktwo}
\textbf{\method} \ours & \best{86.1} & \best{64.6}\\
\bottomrule
\end{tabularx}
\end{table}

\subsection{Real-world experiments}
\label{sec:realworld}
\begin{figure}[t]
\centering
\includegraphics[width=\linewidth]{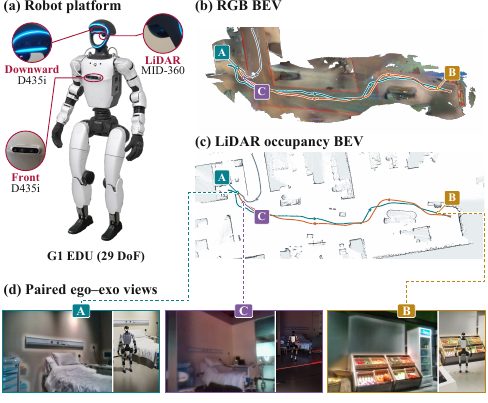}
\caption{\textbf{Real-world platform and memory-reuse task.} (a) The G1 EDU platform and its sensors. (b) RGB BEV and robot trajectory. (c) The corresponding LiDAR occupancy BEV. (d) Paired egocentric (ego, left) and exocentric (exo, right) views of the initial bed visit (A), produce area (B), and return to the same bed (C).}
\label{fig:realworld_platform}
\vspace{-10pt}
\end{figure}

We further evaluate \method's generalization and robustness in the real world. We deploy \method on a 29-DoF Unitree G1 EDU humanoid robot with GPT-6-Astra~\cite{gpt6astra} as the base policy. As shown in \cref{fig:realworld_platform}(a), the platform combines a head-mounted Livox MID-360 LiDAR and a built-in downward-facing Intel RealSense D435i RGB-D camera with an additional forward-facing D435i on the chest.
For navigation, the model points the destination on BEV, which an $\mathrm{A^*}$ planner converts into collision-checked paths. A path-following controller tracks these paths through body-velocity commands issued by the Unitree SDK2 Python G1 locomotion interface.

Starting from a corridor, the robot must first locate and approach a bed to its right. It must then navigate to the fruit-and-vegetable area and finally return to the same bed, reusing memory retained from the initial visit. After the robot leaves the bed, we switch off some of the lights to change the illumination before its return. The robot autonomously completes the entire sequence in closed loop, reaching the fruit-and-vegetable area and subsequently revisiting the original bed under the changed lighting conditions. \cref{fig:realworld_platform}(b)(c) show the RGB reconstruction BEV and the LiDAR occupancy BEV respectively, while \cref{fig:realworld_platform}(d) presents paired ego--exo views of the three stages. This experiment demonstrates successful real-world deployment of \method, connecting VLM-based visual grounding and persistent spatial memory to autonomous execution of a long-horizon task on a physical humanoid.

\section{Analysis}
\label{sec:analysis}

We revisit the three questions for 3D visual harness design in \cref{sec:what_should_agent_see} to \cref{sec:how_should_agent_act}, and analyze capability boundaries of \method in \cref{sec:what_can_harness_address}.

\subsection{What should the agent see?}
\label{sec:what_should_agent_see}
Under partial observability, harness needs to expose proper visual information of the surrounding scene, interaction history, and previous memory to VLM. 
We first examine what spatial observations and contextual cues support VLM decision-making. Results on ProcTHOR (\cref{tab:observation_interface}) show that removing either spatial view or individual BEV context overlays reduces success, with larger losses when a spatial view is removed. 
Therefore, an effective 3D visual harness observation interface should \textit{expose both global and local scene geometry together with the context of the agent's interaction}, providing enough information to the VLM for relating potential destinations to previously visited locations.

\begin{table}[t]
\centering
\caption{\textbf{Observation space study.} SR (\%) on ProcTHOR. Groups (a) and (b) examine what the agent should see through input-view and BEV-overlay removals.}
\label{tab:observation_interface}
\setlength{\tabcolsep}{3pt}
\begin{tabularx}{\linewidth}{>{\raggedright\arraybackslash}Xrr}
\toprule
\rowcolor{rankone}
\multicolumn{3}{l}{\textbf{(a) Input views}}\\
\textbf{Setting} & \textbf{SR} $\uparrow$ & $\boldsymbol{\Delta}$ \textbf{(pp)}\\
\midrule
\rowcolor{ranktwo}
\method \ours & 64.6 & \unreported\\
w/o BEV input & 48.0 & $-16.5$\\
w/o overhead input & 49.6 & $-15.0$\\
\midrule
\rowcolor{rankone}
\multicolumn{3}{l}{\textbf{(b) BEV context overlays}}\\
w/o exploration candidates & 55.9 & $-8.7$\\
w/o past trajectory & 56.7 & $-7.9$\\
w/o previous position & 57.5 & $-7.1$\\
\bottomrule
\end{tabularx}
\end{table}

\subsection{What should the agent remember?}
\label{sec:what_should_agent_remember}
We then explore which information should persist across decisions and goals by removing individual memory components (\cref{tab:memory_design}(a-d)). Removing either policy annotations or grounded evidence groups reduces SR on ProcTHOR, with a larger decline when both are removed. Task-progress, visual-evidence, and scene-memory removals likewise degrade performance on R2R-CE, HM-EQA, and GOAT-Bench, respectively. These components serve complementary roles in successive decisions: task-progress memory tracks what remains to be done, spatial memory locates relevant objects and places, visual evidence memory preserves keyframes from previous observations, and lifelong scene memory carries experience across goals.

We further compare \method with a text-memory configuration, where the agent creates, updates, and deletes persistent natural-language records, instead of keeping and accessing memory visually in \method. It achieves 52.8\% SR on ProcTHOR versus 64.6\% for \method (\cref{tab:memory_design}(e)). In \method, remembered locations remain visually linked to current scene geometry through visualizing on BEV, providing spatial references for subsequent decision-making. 

Therefore, an effective 3D visual harness memory interface should \textit{retain task context and grounded experience across decisions, while preserving reusable scene knowledge across goals}, allowing the VLM to use earlier experience when selecting its next action.

\begin{table}[t]
\centering
\caption{\textbf{Memory design study.} Groups (a)--(d) ablate different memory components; (e) evaluates a text-memory configuration.}
\label{tab:memory_design}
\setlength{\tabcolsep}{3pt}

\begin{tabularx}{\linewidth}{>{\raggedright\arraybackslash}Xrr}
\toprule
\rowcolor{rankone}
\multicolumn{3}{l}{\textbf{(a) Spatial memory (ProcTHOR)}}\\
\textbf{Setting} & \textbf{SR} $\uparrow$ & $\boldsymbol{\Delta}$ \textbf{(pp)}\\
\midrule
\rowcolor{ranktwo}
\method \ours & 64.6 & \unreported\\
w/o grounded evidence groups & 59.1 & $-5.5$\\
w/o policy annotations & 56.7 & $-7.9$\\
w/o both spatial memory sources & 52.8 & $-11.8$\\
\end{tabularx}
\par\smallskip

\begin{tabularx}{\linewidth}{>{\raggedright\arraybackslash}X*{3}{c}}
\rowcolor{rankone}
\multicolumn{4}{l}{\textbf{(b) Task-progress memory (R2R-CE)}}\\
\textbf{Setting} & \textbf{SR} $\uparrow$ & \textbf{SPL} $\uparrow$ & \textbf{nDTW} $\uparrow$\\
\midrule
\rowcolor{ranktwo}
\method \ours & 81.0 & 70.2 & 51.7\\
w/o task progress & 73.0 & 63.2 & 47.9\\
\end{tabularx}
\par\smallskip

\begin{tabularx}{\linewidth}{>{\raggedright\arraybackslash}Xrr}
\rowcolor{rankone}
\multicolumn{3}{l}{\textbf{(c) Visual evidence memory (HM-EQA)}}\\
\textbf{Setting} & \textbf{Acc.} $\uparrow$ & $\boldsymbol{\Delta}$ \textbf{(pp)}\\
\midrule
\rowcolor{ranktwo}
\method \ours & 77.6 & \unreported\\
w/o visual evidence memory & 73.8 & $-3.8$\\
\end{tabularx}
\par\smallskip

\begin{tabularx}{\linewidth}{>{\raggedright\arraybackslash}Xrr}
\rowcolor{rankone}
\multicolumn{3}{l}{\textbf{(d) Lifelong scene memory (GOAT-Bench)}}\\
\textbf{Setting} & \textbf{SR} $\uparrow$ & \textbf{SPL} $\uparrow$\\
\midrule
\rowcolor{ranktwo}
\method \ours & 66.5 & 46.1\\
w/o scene memory & 59.4 & 40.0\\
\end{tabularx}
\par\smallskip

\begin{tabularx}{\linewidth}{>{\raggedright\arraybackslash}Xrr}
\rowcolor{rankone}
\multicolumn{3}{l}{\textbf{(e) Text memory (ProcTHOR)}}\\
\textbf{Setting} & \textbf{SR} $\uparrow$ & $\boldsymbol{\Delta}$ \textbf{(pp)}\\
\midrule
\rowcolor{ranktwo}
\method \ours & 64.6 & \unreported\\
Text memory & 52.8 & $-11.8$\\
\bottomrule
\end{tabularx}
\end{table}

\begin{table}[t]
\centering
\caption{\textbf{Action-space study.} SR (\%) on ProcTHOR. (a) compares views used for policy annotations; (b) compares navigation action space with observation space held fixed.}
\label{tab:action_space}
\setlength{\tabcolsep}{3pt}
\begin{tabularx}{\linewidth}{>{\raggedright\arraybackslash}Xrr}
\toprule
\rowcolor{rankone}
\multicolumn{3}{l}{\textbf{(a) Annotation view}}\\
\textbf{Setting} & \textbf{SR} $\uparrow$ & $\boldsymbol{\Delta}$ \textbf{(pp)}\\
\midrule
\rowcolor{ranktwo}
\method \ours & 64.6 & \unreported\\
BEV & 62.2 & $-2.4$\\
POV & 59.8 & $-4.7$\\
Overhead & 57.5 & $-7.1$\\
\midrule
\rowcolor{rankone}
\multicolumn{3}{l}{\textbf{(b) Navigation}}\\
\textbf{Setting} & \textbf{SR} $\uparrow$ & $\boldsymbol{\Delta}$ \textbf{(pp)}\\
\midrule
\rowcolor{ranktwo}
\method \ours & 64.6 & \unreported\\
BEV pointing & 53.5 & $-11.0$\\
Overhead pointing & 48.8 & $-15.8$\\
POV pointing & 44.1 & $-20.5$\\
Discrete action commands & 22.8 & $-41.7$\\
\bottomrule
\end{tabularx}
\end{table}

\begin{figure*}[t]
\centering
\includegraphics[width=\textwidth]{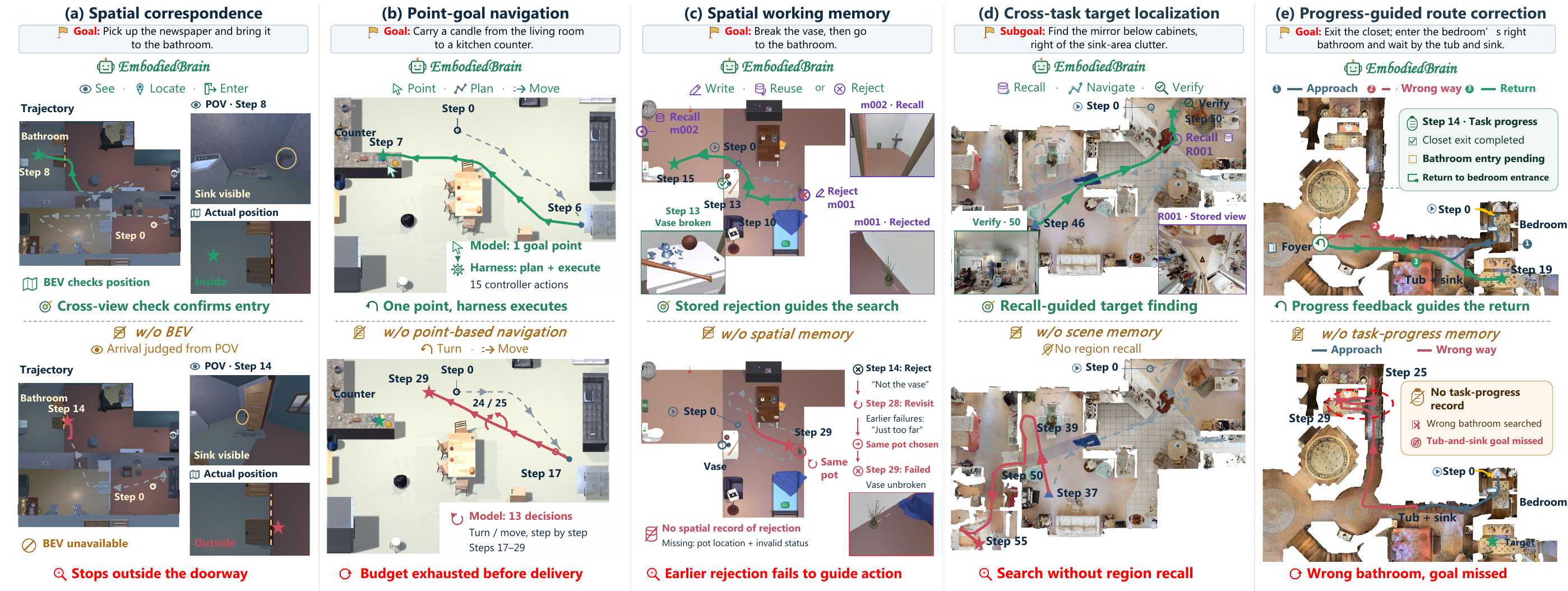}
\caption{\textbf{Harness-enabled capabilities.} Five examples compare \method (top) with ablation (bottom).}
\label{fig:harness_enabled_capabilities}
\end{figure*}

\begin{figure}[t]
\centering
\includegraphics[width=\linewidth]{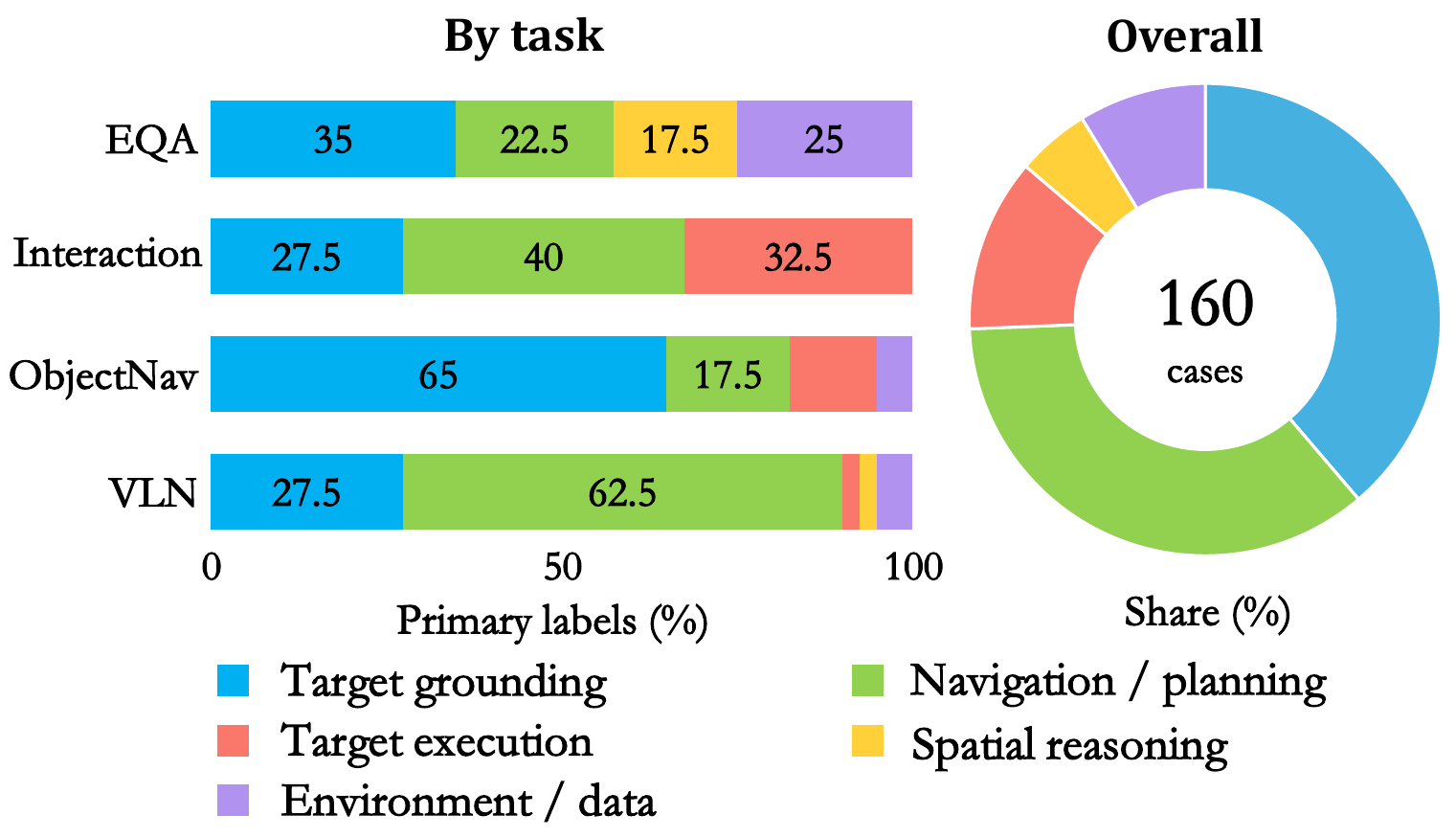}
\caption{\textbf{Error patterns across task groups.} 160 sampled failures from Qwen3.8-27B results, with 40 cases per group.}
\label{fig:error_patterns}
\end{figure}

\subsection{How should the agent act?}
\label{sec:how_should_agent_act}
We next investigate how the joint action interface should be designed. Specifically, we focus on two aspects: how the agent writes policy annotations and how it specifies navigation destinations. For policy annotations, we restrict the source view provided to the VLM for pointing to BEV, POV, or overhead alone, while keeping the navigation action space fixed. As shown in \cref{tab:action_space}(a), BEV alone performs better than the other two, but all three variants perform below \method. For navigation, we fix the observation space and compare \method with discrete commands and single-view pointing variants (\cref{tab:action_space}(b)). Every pointing-based variant outperforms the variant with discrete action commands, while \method performs best with 64.6\% SR, compared with 53.5\% for BEV-only pointing and 22.8\% for discrete commands. 
Therefore, an effective 3D visual harness action interface should \textit{support model-directed spatial annotation and destination selection via pointing in a shared visual workspace}, fully unleashing spatial understanding, reasoning and visual grounding capabilities of VLMs.

\subsection{What can harness address?}
\label{sec:what_can_harness_address}

\paragraph{What does harness enable?}
\label{sec:what_does_harness_enable}

To examine which difficulties the harness helps overcome in long-horizon embodied tasks, we manually check some cases in \cref{sec:what_should_agent_see} to \cref{sec:how_should_agent_act} and summarize five capabilities: spatial correspondence, goal-directed navigation, spatial working memory, cross-task target localization, and progress-guided route correction. The cases in \cref{fig:harness_enabled_capabilities} show how the harness helps existing VLM deal with existing problems.

\paragraph{What remains difficult even with harness?}
\label{sec:what_remains_difficult}

We next examine what kinds of problems still break down even with complete \method harness. We evaluate Qwen3.8-27B on all the benchmarks in \cref{sec:experiments}, sample 40 failure cases for each of four task groups from the results: EQA, Interaction tasks, ObjectNav and VLN, and assign one label per case. Then we distinguish five primary error categories:
\begin{itemize}
\item \textcolor{errorgrounding}{\textbf{Target grounding error.}} The agent fails to identify the correct target, instance, or reference.
\item \textcolor{errorexecution}{\textbf{Target execution error.}} The correct target or answer is identified, but the final action, stop, or submission fails to complete the task.
\item \textcolor{errorspatial}{\textbf{Spatial reasoning error.}} The relevant objects are correctly identified, but their attributes, count, geometry, or spatial relations are inferred incorrectly.
\item \textcolor{errorplanning}{\textbf{Navigation/planning error.}} The goal is understood, but search, route planning, or task decomposition fails to obtain evidence or make progress.
\item \textcolor{errorenvironment}{\textbf{Environment/data error.}} Annotation inconsistencies, environment faults, or problems in mapping, execution infrastructure, or evaluation account for the failure.
\end{itemize}

Target grounding (38.8\%) and navigation/planning (35.6\%) together constitute \textbf{74.4\%} of the sampled failures (\cref{fig:error_patterns}). The dominant category differs by task: grounding represents 65.0\% of navigation failures, whereas planning makes up 62.5\% of VLN failures. Interaction has a substantial final-execution component at 32.5\%, indicating additional difficulty in completing actions after identifying the correct target.

\section{Conclusion}
\label{sec:conclusion}
We presented EmbodiedBrain, a 3D visual harness guided by the  design principle of unleashing existing VLM capabilities for embodied decision-making. Evaluations across five task families and real-world deployment demonstrate its effectiveness. 
To investigate what embodied agents require from a 3D visual harness, we conducted further analysis experiments, and results support our design of organizing observation, memory management, and flexible action selection within a shared spatial workspace, 
providing empirical guidance for 3D visual harness design.
\section*{LLM Usage Statement}
In this work, we used generative AI tools such as Codex for writing portions of the code. Additionally, we used ChatGPT to polish the manuscript. We have reviewed all AI-assisted work. In particular, the authors reviewed the LLM-generated code for correctness and consistency with the intended implementation, and checked the manuscript edits to ensure that they preserved the intended meaning. We take responsibility for the final content of this work, including text, claims or artifacts produced with the aid of generative AI.
{\small
\bibliographystyle{ieeenat_fullname}
\bibliography{bibliography/embodiedbrain}
}
\clearpage
\maketitlesupplementary
\appendix
\setcounter{figure}{0}
\setcounter{table}{0}
\setcounter{equation}{0}
\renewcommand{\thefigure}{S\arabic{figure}}
\renewcommand{\thetable}{S\arabic{table}}
\renewcommand{\theequation}{S\arabic{equation}}
\renewcommand{\theHfigure}{supp.\arabic{figure}}
\renewcommand{\theHtable}{supp.\arabic{table}}
\renewcommand{\theHequation}{supp.\arabic{equation}}
\nocite{pixnav}

\section{More Experiments}
\label{app:more_experiment_results}

\subsection{Detailed results}
\label{app:category_results}

\paragraph{EmbodiedEval.}
\Cref{tab:embodiedeval_categories} separates EmbodiedEval's original five task categories. With Qwen3.8-27B, \method achieves 64.21\% SR on attribute QA, 57.69\% on spatial QA, and 58.62\% on navigation. It completes 39 of 89 object-interaction tasks and 15 of 34 social-interaction tasks, yielding 43.82\% and 44.12\% SR, respectively, surpassing all the listed baselines in all five categories.

\begin{table*}[h]
\centering
\captionsetup{singlelinecheck=false}
\caption{\textbf{EmbodiedEval by task category.} SR (\%) across the five original categories and Overall. Best in bold and second best underlined.}
\label{tab:embodiedeval_categories}
\setlength{\tabcolsep}{3pt}
\begin{tabularx}{\textwidth}{@{}L{0.26\textwidth}*{3}{>{\centering\arraybackslash}X}*{2}{>{\centering\arraybackslash}p{0.14\textwidth}}>{\centering\arraybackslash}X@{}}
\toprule
\textbf{Method} & \mbox{\textbf{Attribute QA}} & \mbox{\textbf{Spatial QA}} & \textbf{Navigation} & \mbox{\textbf{Object Interaction}} & \mbox{\textbf{Social Interaction}} & \textbf{Overall}\\
\midrule
GPT-4o~\cite{embodiedeval} & \second{35.79} & \second{32.69} & \second{31.03} & \second{10.11} & \second{11.76} & \second{25.00}\\
Gemini-Pro-1.5~\cite{embodiedeval} & 27.37 & 9.62 & 17.24 & 4.49 & 5.88 & 14.33\\
SEEA-R1~\cite{seea_r1} & 30.85 & \second{32.69} & 21.05 & 4.60 & 6.25 & 19.88\\
\midrule
\rowcolor{ranktwo}
\textbf{\method} \ours & \best{64.21} & \best{57.69} & \best{58.62} & \best{43.82} & \best{44.12} & \best{54.57}\\
\bottomrule
\end{tabularx}
\par\smallskip
\end{table*}

\paragraph{HM-EQA}
\Cref{tab:hmeqa_categories} breaks HM-EQA accuracy down by question type for the Qwen3.8-Flash-Next configuration. \method improves on the strongest listed baseline in counting, existence, state, and location. Counting remains our lowest-scoring category at 71.3\%, despite a 13.4-point improvement over GraphEQA.

\begin{table*}[h]
\centering
\captionsetup{singlelinecheck=false}
\caption{\textbf{HM-EQA by question type.} Accuracy (\%) by question type and overall. Baselines and $\dagger$-marked reproductions follow GraphEQA~\cite{grapheqa}. $\ddagger$ marks Explore-EQA scores from the GraphEQA comparison. Best in bold and second best underlined.}
\label{tab:hmeqa_categories}
\setlength{\tabcolsep}{3pt}
\begin{tabularx}{\textwidth}{@{}L{0.26\textwidth}*{3}{>{\centering\arraybackslash}X}*{2}{>{\centering\arraybackslash}p{0.14\textwidth}}>{\centering\arraybackslash}X@{}}
\toprule
\textbf{Method} & \textbf{Identification} & \textbf{Counting} & \textbf{Existence} & \textbf{State} & \textbf{Location} & \textbf{Overall}\\
\midrule
Explore-EQA$^{\ddagger}$~\cite{grapheqa} & 59.2 & 46.2 & 56.5 & 46.5 & 47.7 & 51.7\\
Explore-EQA (GPT-4o)$^{\dagger}$~\cite{grapheqa} & 32.5 & 44.2 & 56.4 & 42.3 & 40.8 & 46.4\\
SayPlanVision$^{\dagger}$~\cite{grapheqa} & 75.0 & 44.4 & 63.3 & 43.4 & 56.0 & 54.8\\
GraphEQA~\cite{grapheqa} & \best{77.8} & \second{57.9} & \second{69.0} & \second{65.2} & \second{64.0} & \second{63.5}\\
\midrule
\rowcolor{ranktwo}
\textbf{\method} \ours & \second{76.5} & \best{71.3} & \best{83.3} & \best{76.2} & \best{78.5} & \best{77.6}\\
\bottomrule
\end{tabularx}
\par\smallskip
\end{table*}

\paragraph{VLN.}
\Cref{tab:vln_trajectory_metrics} extends the main VLN experiment with six metrics for both R2R-CE and RxR-CE: success rate (SR), oracle success rate (OSR), success weighted by path length (SPL), navigation error (NE), normalized dynamic time warping (nDTW), and success-weighted nDTW (sDTW). Metric definitions and our evaluation setup are given in \cref{sec:settings-vln}.

\subsection{Extended VLN Comparisons}
\label{app:vln_full_results}

\begin{table*}[!t]
\centering
\captionsetup{singlelinecheck=false}
\caption{\textbf{Extended VLN-CE comparison.} Six metrics on R2R-CE and RxR-CE. NE is in meters and lower is better. Other metrics are percentages and higher is better. Superscripts M and P indicate monocular and panoramic RGB, respectively. Full follows each source's full val-unseen protocol. ON-100 denotes the Open-Nav 100-episode R2R cohort, and ON-50 denotes 50 of those tasks. ON/Full and ON/R100 pair ON-100 with full val-unseen and 100 randomly sampled English-guide RxR tasks, respectively. The best and second-best reported values are bold and underlined. A dash indicates an unreported value or group.}
\label{tab:vln_trajectory_metrics}
\setlength{\tabcolsep}{1.1pt}
\begin{tabularx}{\textwidth}{@{}L{0.250\textwidth}>{\centering\arraybackslash}p{0.074\textwidth}*{6}{>{\centering\arraybackslash}X}@{\hspace{5pt}}*{6}{>{\centering\arraybackslash}X}@{}}
\toprule
\multirow{2}{*}{\textbf{Method / backbone}} & \multirow{2}{*}{\textbf{Split}} & \multicolumn{6}{c}{\textbf{R2R-CE}} & \multicolumn{6}{c}{\textbf{RxR-CE}}\\
\cmidrule(lr){3-8}\cmidrule(lr){9-14}
& & SR & SPL & OSR & NE & nDTW & sDTW & SR & SPL & OSR & NE & nDTW & sDTW\\
\midrule
\multicolumn{14}{@{}l}{\textit{Training-based methods}}\\
NaVid~\cite{navid} & Full & 41.9 & 36.5 & 49.2 & 5.72 & \unreported & \unreported & 45.7 & 38.2 & \unreported & 5.72 & \unreported & \unreported\\
Uni-NaVid~\cite{uninavid} & Full & 47.0 & 42.7 & 53.3 & 5.58 & \unreported & \unreported & 48.7 & 40.9 & \unreported & 6.24 & \unreported & \unreported\\
NaVILA~\cite{navila} & Full & 54.0 & 49.0 & 62.5 & 5.22 & \unreported & \unreported & 49.3 & 44.0 & \unreported & 6.77 & 58.8 & \unreported\\
JanusVLN (Extra)~\cite{janusvln} & Full & 60.5 & 56.8 & 65.2 & 4.78 & \unreported & \unreported & 56.2 & 47.5 & \unreported & 6.06 & 62.1 & \unreported\\
StreamVLN~\cite{streamvln} & Full & 56.9 & 51.9 & 64.2 & 4.98 & \unreported & \unreported & 52.9 & 46.0 & \unreported & 6.22 & 61.9 & \unreported\\
DualVLN~\cite{dualvln} & Full & 64.3 & 58.5 & 70.7 & 4.05 & \unreported & \unreported & 61.4 & 51.8 & \unreported & 4.58 & 70.0 & \unreported\\
InternVLA-N1~\cite{internvlan1} & Full & 58.2 & 54.0 & 63.3 & 4.83 & \unreported & \unreported & 53.5 & 46.1 & \unreported & 5.91 & 65.3 & \unreported\\
Qwen-RobotNav-4B\textsuperscript{M}~\cite{qwenrobotnav} & Full & 66.9 & 60.5 & 73.6 & 4.22 & \unreported & \unreported & 71.3 & 61.5 & \unreported & 4.15 & 68.6 & \unreported\\
Qwen-RobotNav-8B\textsuperscript{M}~\cite{qwenrobotnav} & Full & 65.7 & 59.6 & 72.7 & 4.36 & \unreported & \unreported & 73.4 & 63.5 & \unreported & 4.16 & 69.9 & \unreported\\
NavFoM~\cite{navfom} & Full & 61.7 & 55.3 & 72.1 & 4.61 & \unreported & \unreported & 64.4 & 56.2 & \unreported & 4.74 & 65.8 & \unreported\\
ABot-N0~\cite{abotn0} & Full & 66.4 & 63.9 & 70.8 & 3.78 & \unreported & \unreported & 69.3 & 60.0 & \unreported & 3.83 & \unreported & \unreported\\
OmniNav~\cite{omninav} & Full & 69.5 & 66.1 & 74.6 & 3.74 & \unreported & \unreported & 73.6 & 62.0 & \unreported & 3.77 & \unreported & \unreported\\
AstraNav-World~\cite{astranavworld} & Full & 67.9 & 65.4 & 73.9 & 3.86 & \unreported & \unreported & 72.9 & 61.5 & \unreported & 3.82 & \unreported & \unreported\\
Qwen-RobotNav-4B\textsuperscript{P}~\cite{qwenrobotnav} & Full & 69.5 & 63.6 & 77.2 & 3.80 & \unreported & \unreported & 75.2 & \second{65.0} & \unreported & 3.80 & 71.9 & \unreported\\
Qwen-RobotNav-8B\textsuperscript{P}~\cite{qwenrobotnav} & Full & 72.1 & \second{66.6} & \second{78.5} & \second{3.53} & \unreported & \unreported & \second{76.5} & \best{65.7} & \unreported & \second{3.58} & \second{72.5} & \unreported\\
\midrule
\multicolumn{14}{@{}l}{\textit{Zero-shot methods}}\\
Open-Nav~\cite{opennav}: GPT-4 & ON-100 & 19.0 & 16.1 & 23.0 & 6.70 & 45.8 & \unreported & \multicolumn{6}{c@{}}{\unreported}\\
Open-Nav~\cite{opennav}: Gemini-3-Flash & ON-100 & 32.0 & 27.7 & 42.0 & 6.42 & 54.8 & \unreported & \multicolumn{6}{c@{}}{\unreported}\\
Three-Step Nav~\cite{threestepnav}: GPT-5 & ON-100 & 34.0 & 29.1 & 39.0 & 5.87 & 57.7 & \unreported & \multicolumn{6}{c@{}}{\unreported}\\
EvoNav~\cite{evonav}: Gemini-2.5-Pro & ON-100 & 43.0 & 37.8 & 51.0 & 5.04 & 62.4 & \unreported & \multicolumn{6}{c@{}}{\unreported}\\
SmartWay~\cite{smartway}: GPT-5.5 & ON-100 & 44.0 & 35.0 & 60.0 & 5.16 & 58.6 & \unreported & \multicolumn{6}{c@{}}{\unreported}\\
AgenticNav~\cite{agenticnav}: Gemini-2.5-Pro & ON-100 & 49.0 & 33.2 & 63.0 & 5.91 & 48.7 & \unreported & \multicolumn{6}{c@{}}{\unreported}\\
AgenticNav~\cite{agenticnav}: GPT-5.5 & ON-100 & 55.0 & 48.4 & 65.0 & 5.19 & \second{63.4} & \unreported & \multicolumn{6}{c@{}}{\unreported}\\
GPT-6-Astra workflow~\cite{astra_vln}:\newline medium & ON-50 & 52.0 & 48.9 & 62.0 & 4.34 & \best{70.8} & \unreported & \multicolumn{6}{c@{}}{\unreported}\\
HarnessVLN~\cite{harnessvln}: GPT-5.5 & ON/Full & 58.0 & 35.8 & \unreported & \unreported & \unreported & \unreported & 51.4 & 38.1 & \unreported & \unreported & \unreported & \unreported\\
Minimal agent~\cite{agents_control}: fable-5 & ON-100 & \second{78.0} & 65.3 & \unreported & \unreported & \unreported & \unreported & \multicolumn{6}{c@{}}{\unreported}\\
\midrule
\rowcolor{ranktwo}
\textbf{\method} \ours & ON/R100 & \best{81.0} & \best{70.2} & \best{82.0} & \best{2.63} & 51.7 & \best{46.8} & \best{81.0} & 62.8 & \best{83.0} & \best{3.07} & \best{77.4} & \best{69.4}\\
\bottomrule
\end{tabularx}
\par\smallskip
\end{table*}

For a matched comparison, we evaluate five trained baselines in \cref{tab:vln_subset_trained} on the same episodes as \method, using Open-Nav's ON-100 subset for R2R-CE and the 100 sampled English-guide episodes described in \cref{sec:vln_results} for RxR-CE. For HarnessVLN~\cite{harnessvln}, we use its published ON-100 result and evaluate it on our RxR-CE subset.

HarnessVLN achieves 58.0\% SR and 35.8\% SPL on R2R-CE and 52.0\% SR and 30.3\% SPL on RxR-CE. Among the training-based baselines in the table, OmniNav achieves the highest SR and SPL on both matched subsets, while \method reaches 81.0\% SR on both benchmarks, exceeding OmniNav by 12.0 percentage points on R2R-CE and 14.0 points on RxR-CE. Its SPL reaches 70.2\% and 62.8\%, respectively, improving on OmniNav by 6.4 and 7.5 points. On these tasks, a general-purpose VLM equipped with our 3D visual harness succeeds more often than the navigation-trained baselines in the table and retains its lead after accounting for path length.

\begin{table}[h]
\centering
\captionsetup{singlelinecheck=false}
\caption{\textbf{Navigation on 100-episode subsets.} SR and SPL (\%) on R2R-CE ON-100 and the sampled English-guide RxR-CE episodes.}
\label{tab:vln_subset_trained}
\setlength{\tabcolsep}{3pt}
\begin{tabularx}{\linewidth}{@{}L{0.40\linewidth}*{4}{>{\centering\arraybackslash}X}@{}}
\toprule
\multirow{2}{*}{\textbf{Method}} & \multicolumn{2}{c}{\textbf{R2R-CE}} & \multicolumn{2}{c}{\textbf{RxR-CE}} \\
\cmidrule(lr){2-3}\cmidrule(lr){4-5}
& SR $\uparrow$ & SPL $\uparrow$ & SR $\uparrow$ & SPL $\uparrow$ \\
\midrule
\multicolumn{5}{@{}l}{\textit{Training-based methods}}\\
NaVILA~\cite{navila} & 50.0 & 45.2 & 52.0 & 44.6 \\
JanusVLN (Extra)~\cite{janusvln} & 57.0 & 54.0 & 56.0 & 44.4 \\
Uni-NaVid~\cite{uninavid} & 59.0 & 54.3 & 59.0 & 46.7 \\
DualVLN~\cite{dualvln} & 65.0 & 58.0 & 56.0 & 48.1 \\
OmniNav~\cite{omninav} & \second{69.0} & \second{63.8} & \second{67.0} & \second{55.3} \\
\midrule
\multicolumn{5}{@{}l}{\textit{Zero-shot methods}}\\
HarnessVLN~\cite{harnessvln} & 58.0 & 35.8 & 52.0 & 30.3 \\
\rowcolor{ranktwo}
\textbf{\method} \ours & \best{81.0} & \best{70.2} & \best{81.0} & \best{62.8} \\
\bottomrule
\end{tabularx}
\end{table}

\subsection{Does \method{} Improve Task Success Across Model Backbones?}
\label{sec:matched_models}
In \Cref{sec:interaction_results}, \method improves the performance of Qwen3.8-27B on ProcTHOR interaction tasks compared to SpatialWorld native harness. To examine whether this gain extends to other VLMs, we evaluate eight backbones on the same ProcTHOR subset with both harnesses. All eight achieve higher success rates with \method{}, with gains ranging from 5.5 to 62.2 percentage points (\cref{tab:matched_models}). The gains span backbones with widely different native-harness success rates, suggesting that a shared 3D visual harness can help different VLMs use their existing capabilities for embodied decision-making.

\begin{table*}[t]
\centering
\captionsetup{singlelinecheck=false}
\caption{\textbf{SpatialWorld model comparison on ProcTHOR.} Naive denotes SpatialWorld's native harness. Steps averages high-level decisions across successful and failed tasks that did not terminate due to protocol errors.}
\label{tab:matched_models}
\setlength{\tabcolsep}{4pt}
\definecolor{s5delta}{HTML}{216B40}
\definecolor{s5increase}{HTML}{A52C32}
\newcommand{\sFiveDelta}[3][s5delta]{\makebox[2.8em][r]{#2}\hspace{0.35em}\makebox[4.4em][l]{\textcolor{#1}{\ensuremath{(#3)}}}}
\begin{tabularx}{\textwidth}{@{\hspace{3pt}}L{0.28\textwidth}*{2}{>{\centering\arraybackslash}X}@{\hspace{12pt}}*{2}{>{\centering\arraybackslash}X}@{\hspace{3pt}}}
\toprule
\multirow{2}{*}{\textbf{Model}} & \multicolumn{2}{c}{\textbf{Naive}} & \multicolumn{2}{c}{\cellcolor{ranktwo}\textbf{EmbodiedBrain}} \\
\cmidrule(lr){2-3}\cmidrule(lr){4-5}
& SR $\uparrow$ & Steps $\downarrow$ & SR $\uparrow$ & Steps $\downarrow$ \\
\midrule
Qwen3.5-9B & 0.00 & 24.19 & \sFiveDelta{34.65}{+34.65} & \sFiveDelta{20.17}{-4.02} \\
InternVL3.5-38B-Instruct & 0.00 & 24.35 & \sFiveDelta{5.51}{+5.51} & \sFiveDelta[s5increase]{24.69}{+0.34} \\
Qwen3.6-35B-A3B & 0.79 & 22.06 & \sFiveDelta{39.37}{+38.58} & \sFiveDelta{19.62}{-2.44} \\
Qwen3.8-27B & 2.36 & 35.62 & \sFiveDelta{64.57}{+62.21} & \sFiveDelta{15.84}{-19.78} \\
GLM-4.6V & 3.15 & 32.66 & \sFiveDelta{16.54}{+13.39} & \sFiveDelta{26.51}{-6.15} \\
Qwen3.8-Flash-Next & 9.45 & 40.42 & \sFiveDelta{68.50}{+59.05} & \sFiveDelta{17.04}{-23.38} \\
MiMo-V2.6-Flash-RL & 20.47 & 46.84 & \sFiveDelta{46.46}{+25.99} & \sFiveDelta{21.28}{-25.56} \\
GPT-6.1-Sol & 57.48 & 31.13 & \sFiveDelta{75.59}{+18.11} & \sFiveDelta{14.20}{-16.93} \\
\bottomrule
\end{tabularx}
\end{table*}

\section{Method Details}
\label{app:method_details}

This section specifies the action, reconstruction, and memory mechanisms of the harness. The adapters share an update-and-act loop while retaining benchmark-specific skills and completion rules. Evaluation settings are specified in Experiment Details.

Spatial memory consists of policy annotations and grounded evidence. Each grounded evidence group $e_j=(\rho_j,X_j,d_j)$ links an observing camera pose, an estimated object location, and its semantic description. Multiple entries can share an observing camera pose.

\subsection{Action interfaces}
\label{app:action_interfaces}
The policy receives the task instruction, current images, enabled memory readouts, and recent execution feedback. It specifies an environment action and any supported memory edit actions. \Cref{tab:adapter_actions} lists the environment actions and how the adapter executes them. The available memory edit actions follow the interfaces in \cref{app:memory_formats,app:lifelong_implementation}.

\begin{table*}[t]
\centering
\captionsetup{singlelinecheck=false}
\caption{\textbf{Action interfaces by benchmark.}}
\label{tab:adapter_actions}
\begin{tabularx}{\textwidth}{@{}L{0.23\textwidth}L{0.19\textwidth}>{\raggedright\arraybackslash}XL{0.17\textwidth}L{0.17\textwidth}@{}}
\toprule
\textbf{Benchmark / subset} & \textbf{Navigation} & \textbf{Other actions} & \textbf{Completion} & \textbf{Controller} \\
\midrule
\rowcolor{ranktwo}
\multicolumn{5}{@{}l}{\textbf{Vision-and-language navigation}}\\
R2R-CE / RxR-CE & BEV or overhead point & Viewpoint changes & \texttt{stop} & Point-navigation controller \\
\midrule
\rowcolor{ranktwo}
\multicolumn{5}{@{}l}{\textbf{Object-goal navigation and lifelong tasks}}\\
HM3D-v2 ObjectNav & BEV or overhead point & Viewpoint changes & \texttt{stop} & \multirow[t]{3}{=}{Point-navigation controller} \\
MP3D ObjectNav & BEV or overhead point & Viewpoint changes & \texttt{stop} &  \\
GOAT-Bench & BEV or overhead point & Viewpoint changes & \texttt{subtask\_stop}, then \texttt{stop} after all goals &  \\
\midrule
\rowcolor{ranktwo}
\multicolumn{5}{@{}l}{\textbf{Embodied question answering}}\\
HM-EQA & BEV or overhead point & Viewpoint changes & \texttt{answer} (A--D) & \multirow[t]{3}{=}{Point-navigation controller} \\
MT-HM3D & BEV or overhead point & Viewpoint changes, stair traversal & \texttt{answer} (A--D) &  \\
EXPRESS-Bench & BEV or overhead point & Viewpoint changes, stair traversal & \texttt{answer} (free text) &  \\
\midrule
\rowcolor{ranktwo}
\multicolumn{5}{@{}l}{\textbf{Interaction and EmbodiedEval}}\\
SpatialWorld AI2-THOR / ProcTHOR & BEV or overhead point & Viewpoint changes, object skills & \texttt{done} & \multirow[t]{4}{=}{Point-navigation controller} \\
EmbodiedEval Navigation & BEV or overhead point & Viewpoint changes & Task-defined &  \\
EmbodiedEval EQA & BEV or overhead point & Viewpoint changes & Submit a listed answer &  \\
EmbodiedEval Interaction & BEV or overhead point & Viewpoint changes, interaction options & Task-defined &  \\
\bottomrule
\end{tabularx}
\end{table*}

A point-navigation action carries a two-element \texttt{pixel} coordinate in the named image. For example, when the policy selects free floor in the occupancy BEV, the harness maps the pixel to a world target. The navigation controller checks traversability and executes the planned route. The RGB reconstruction BEV supplies visual context.

SpatialWorld maps pickup, placement, opening, and closing to simulator skills through \texttt{spatialworld\_action}. The adapter also supports standalone memory edits. EmbodiedEval validates answer and interaction indices against the options in the current observation.

The schematic JSON examples use the memory component names from the paper. The SpatialWorld example below pairs a navigation action with a task-progress memory edit.

\noindent\begin{minipage}{\linewidth}
\footnotesize
\begin{verbatim}
{
  "type": "navigate_bev_point",
  "pixel": [500, 300],
  "reason": "Inspect the adjoining room.",
  "memory_edits": {
    "task_progress_memory": {
      "current_subgoal": "Inspect the room."
    }
  }
}
\end{verbatim}
\end{minipage}\par

\paragraph{Validation and feedback.}
The action schema restricts operations to the task state, available image sources, and writable IDs. The harness validates outputs before execution. Omitted optional edits leave the corresponding memory unchanged.

\subsection{Reconstruction and execution}
\label{app:reconstruction}
Point coordinates use the inclusive normalized range $[0,1000]^2$. For an image of width $W$ and height $H$, $(u,v)$ maps to $(u(W-1)/1000,v(H-1)/1000)$. Coordinates belong to the named source image: a point in the POV cannot be copied to the overhead or BEV image without a geometric transformation. With valid axial depth $D(p)$, camera intrinsics $K$, and camera-to-world pose $(R,t)$, an image pixel $p=(p_x,p_y)$ is back-projected as
\begin{equation}
X=R\bigl(D(p)K^{-1}[p_x,p_y,1]^\top\bigr)+t.
\label{eq:app-backprojection}
\end{equation}
The implementation accounts for each simulator's camera-axis convention. Points in the reconstruction-rendered overhead view are mapped back to the geometric map using the rendering geometry; BEV points use the inverse of the current display transform. Thus a stored world anchor retains its location as the agent and map display move, while its projected image coordinates change.

Valid posed RGB-D observations are fused into colored 3D points and floor/obstacle evidence. The mapper uses a bounded voxelized point store, replacing samples at occupied voxels and optionally cleaning stale geometry. Height-filtered evidence updates occupancy. The RGB reconstruction BEV projects reconstructed points with height-based visibility and ceiling filtering. The POV image provides the current camera observation. The overhead image, occupancy BEV, and RGB reconstruction BEV in the standard interface are rendered from the 3D geometric map.

Memory edit actions use the source observation and precede environment execution, including image-based policy annotations and visual evidence entries. The navigation controller then checks or projects the selected target onto its traversable geometry, constructs a route, and executes simulator actions. Execution returns the actual reached pose and intermediate RGB-D observations, rather than assuming that the selected endpoint was reached. Failed or truncated movement is reported from the reached state.

Between policy decisions $t$ and $t+1$, let $\tilde o_{t,k}$ denote the accepted posed RGB-D measurements from execution substeps and subsequent boundary sensing. The mapper updates the 3D geometric map as
\begin{equation}
\mathcal{C}_{t,k+1}=\mathcal{F}_{\mathrm{geom}}(\mathcal{C}_{t,k},\tilde o_{t,k+1}),
\label{eq:substep_reconstruction}
\end{equation}
with $\mathcal{C}_{t,0}=\mathcal{C}_t$ and $\mathcal{C}_{t+1}=\mathcal{C}_{t,K_t}$ for $K_t$ accepted measurements. These raw sensor packets are distinct from the partial model-visible observation $o_t$. Colored geometry and occupancy are updated before the next policy decision, with auxiliary grounding applied where enabled. Retained source frames, calibration, display transforms, and execution context belong to the harness state defined in \cref{sec:preliminaries}; the model receives their partial visual and textual readout.

\paragraph{Grounding and floor state.}
Where enabled, an auxiliary visual-grounding call processes observed route images and creates grounded evidence groups, each linking an observing viewpoint to an estimated object location and its semantic description. The grounding model uses the same VLM backbone as the policy $\pi_\theta$. The grounding module maintains their geometry separately from policy annotations, and the grounded evidence remains fallible. In multi-floor environments, the active map and spatial memory follow the agent's floor while retaining their semantic annotations.

\subsection{Task-local memory}
\label{app:memory_formats}
Task-local memory contains task-progress memory, spatial memory, and enabled visual evidence memory. The policy specifies memory edits using current images or listed records; the harness maintains geometry and provenance and exposes descriptions and current projections.

\paragraph{Task-progress memory.}
Task-progress memory $\mathcal{M}_t^{\mathrm{prog}}$ retains the current subgoal, recorded findings, checked areas, missing evidence, and counting notes. We apply incremental patches; omitted fields retain their values. Interaction adapters also track completed operations, remaining work, unhelpful attempts, and held-object state, distinguishing execution-confirmed state from the proposed plan. R2R and RxR require progress text with a plan and completed/current/next stages, revised against the executed route.

\paragraph{Policy annotations.}
An addition names a supported source view and normalized point, with a label and adapter-supported category or confidence. The harness resolves the point using that source observation, assigns an annotation ID, and stores its world anchor. Current-image sources include \texttt{current\_pov}, \texttt{current\_overhead}, and \texttt{current\_bev} where supported. An \texttt{initial\_scan} addition must identify an actually attached scan image through its exact \texttt{view\_label}; this source is unavailable after that scan leaves the current input. The following payload illustrates a policy annotation edit.
\par\noindent\begin{minipage}{\linewidth}\footnotesize
\begin{verbatim}
{
  "spatial_memory": {
    "add": [{
      "source": "current_bev",
      "pixel": [420, 610],
      "label": "room entrance",
      "category": "route_landmark",
      "confidence": 0.8
    }]
  }
}
\end{verbatim}
\end{minipage}\par

\paragraph{Semantic updates and ownership.}
The adapter exposes \texttt{spatial\_memory.add}, \texttt{update}, and \texttt{delete\_ids}. Updates refer to currently writable policy annotations or exposed object-location hypotheses within grounded evidence groups and change semantic fields: label, status, confidence, or reason. They do not move an anchor, change its source/floor, or manufacture grounding evidence. New policy annotations start unverified. Available statuses are \texttt{todo}, \texttt{unverified}, \texttt{verified}, \texttt{checked\_empty}, \texttt{completed}, and \texttt{invalid}. These encode model judgements; \texttt{verified} is not evaluator-certified task success. The following example illustrates a memory edit that marks the writable record \texttt{m001} as verified.
\par\noindent\begin{minipage}{\linewidth}\footnotesize
\begin{verbatim}
{
  "spatial_memory": {
    "update": [{
      "id": "m001",
      "status": "verified",
      "confidence": 0.95,
      "reason": "Current POV confirms it."
    }]
  }
}
\end{verbatim}
\end{minipage}\par

Setting \texttt{status=invalid} excludes a disproved record from the active context without erasing its stored history; \texttt{delete\_ids} removes a record when the adapter permits it. Updates precede deletions and additions, allowing a disproved hypothesis to be replaced using a new observation. Target-verification rules can restrict deletion of goal-associated records.

A grounded evidence group's observing-viewpoint component retains the observer position and viewing direction, linked to the group's object-location hypothesis. The policy reads the group's description, available projections, and association without setting its world geometry. Observing-viewpoint components are not directly editable by the policy. Editing a writable object-location hypothesis changes its semantic annotation, while invalidating or deleting it also triggers cleanup of its associated observing-viewpoint component. Models use the spatial-memory interface for these operations.

\paragraph{Visual evidence memory.}
Where enabled, visual evidence memory $\mathcal{M}_t^{\mathrm{ve}}$ holds up to $B_{\mathrm{ve}}=3$ entries. The policy populates it by saving the current POV image or a named supported view with a label and reason, and deletes currently retained IDs. The following payload illustrates saving the current POV as a visual evidence memory entry.
\par\noindent\begin{minipage}{\linewidth}\footnotesize
\begin{verbatim}
{
  "visual_evidence_memory": {
    "save_current_pov": {
      "label": "table contents",
      "reason": "Retain counting evidence."
    }
  }
}
\end{verbatim}
\end{minipage}\par
Stored entries retain source pose and image evidence. Habitat visual evidence memory entries pair the selected POV image with the corresponding overhead image; interaction adapters retain their supported visual sources. New IDs become usable after the runtime returns them in a subsequent observation. Saving is bounded by the adapter's configured pool, with deletion required to free a full pool. Unavailable save sources and noncurrent deletion IDs are rejected rather than treated as successful annotations.

\subsection{Lifelong scene memory}
\label{app:lifelong_implementation}
The GOAT-Bench adapter separates lifelong scene memory $\mathcal{M}_t^{\mathrm{scene}}$ from the task-local memory for the current goal. Region records retain geometric membership, semantic descriptions, and representative images; object records retain grounded positions, descriptions, attributes, and region associations. Auxiliary visual grounding proposes observations, and the harness validates their evidence and geometric associations before updating the bank. The main action policy does not directly redefine region geometry or arbitrarily create persistent object identities.

Regions are derived from observed free space. Erosion separates narrow connections to form interior seeds; connected seeds are expanded within free-space support, and overlap with existing regions stabilizes their identities as the map grows. Visual observations supply semantic descriptions and object-association proposals. Geometric and region-consistency checks constrain those proposals; uncertain associations use a conservative matching fallback rather than automatically creating a confirmed duplicate.

An object record becomes eligible for the policy input with recorded target-view evidence or at least two independent observations, excluding conflicting category evidence. Independence requires at least $0.35$\,m translation from stored observation positions; turning in place alone does not provide another independent observation. Target-view evidence is assessed from observations by the grounding pipeline, not supplied as evaluator truth. Persistent object records expose semantic descriptions and current projections, while the harness retains their world positions and provenance internally.

\paragraph{Goal transition and retrieval.}
At \texttt{subtask\_stop}, the adapter preserves the 3D geometric map and lifelong scene memory while resetting task-progress memory, spatial memory (policy annotations and grounded evidence groups), enabled visual evidence memory, and the previous working set. The displayed reconstruction layer and past-trajectory overlay restart for the new goal; this visual reset does not erase the persistent mapper or scene records. The memory selector receives the new goal, eligible object/region scene records, historical map/reconstruction views, and up to two representative images per region. It selects a small working set and a navigation brief. A shortened illustrative selection is
\par\noindent\begin{minipage}{\linewidth}\footnotesize
\begin{verbatim}
{
  "object_ids": ["O0024"],
  "region_ids": [],
  "navigation_brief": {
    "current_subgoal": "Approach O0024.",
    "plan": "Approach and check current POV.",
    "reason": "It locates the cooking wall."
  }
}
\end{verbatim}
\end{minipage}\par

The implementation retains at most two explicit selections, prioritizing valid objects before regions, and automatically includes each selected object's region. Selected regions can supply temporary safe-floor anchors for navigation. Brief fields are bounded to 800 characters. The policy receives this working context and the records' current projections during subsequent decisions. Historical evidence selects where to search; it does not prove current target visibility or successful completion. Image goals require instance verification in the current view after approach.

A valid empty selection leads to exploration. The retrieval call receives no evaluator feedback about the previous subtask. Scene knowledge persists across goals within the episode; each new episode initializes independent geometry and memory, including when it revisits the same physical scene.

\section{Experiment Details}
\label{app:experiment_details}

\subsection{Benchmarks}
\label{app:benchmarks}

\paragraph{SpatialWorld.}
SpatialWorld~\cite{spatialworld} tests interactive spatial reasoning through a common action interface spanning eight simulators. Its AI2-THOR subset uses interactive indoor scenes grounded in household layouts~\cite{ai2thor}. Its ProcTHOR subset uses procedurally generated homes with varied floor plans and furnishings~\cite{procthor}. The instructions range from reaching a named location to finding, moving, or changing the state of objects in a specified arrangement. Each task pairs its instruction with a verifier that checks the required conditions in the simulator.

\paragraph{EmbodiedEval.}
EmbodiedEval~\cite{embodiedeval} defines 328 tasks across 125 scenes in the LEGENT simulator. It covers attribute question answering, spatial question answering, navigation, object interaction, and social interaction. The agent gathers evidence by moving through the scene and can select from task-specific interaction or answer options. Tasks range from inspecting a spatial relation to delivering an object to a person.

\paragraph{MP3D ObjectNav.}
Matterport3D~\cite{mp3d} provides textured reconstructions and RGB-D panoramas of real indoor spaces, together with semantic object annotations. The object-goal navigation task starts the agent at a pose in one of these scenes and specifies a target category. The agent must explore the environment, find an instance of that category, and reach a valid viewing position.

\paragraph{HM3D ObjectNav.}
Habitat-Matterport 3D (HM3D)~\cite{hm3d} contains building-scale textured reconstructions of real interiors, including multi-floor residences and commercial spaces. In the HM3D ObjectNav task, semantic annotations identify instances of the requested category and support selecting eligible viewpoints. The larger scenes require agents to search across rooms for a reachable target.

\paragraph{GOAT-Bench.}
GOAT-Bench~\cite{goatbench} poses a sequence of object-navigation subtasks in a single environment. Goals can be specified by a category name, a description referring to a particular object, or an image of the target instance. The agent receives the next goal after completing or exhausting the current subtask and continues in the same scene, where earlier observations can inform later searches.

\paragraph{HM-EQA.}
HM-EQA~\cite{exploreeqa} pairs multiple-choice questions with HM3D indoor scenes. Its questions cover object identification, counting, existence, state, and location. They are grounded in household scenarios, such as checking whether an item was left in a particular place, and require the agent to explore for visual evidence.

\paragraph{MT-HM3D.}
MT-HM3D~\cite{memoryeqa} extends HM3D-based question answering to multiple target objects and regions. Questions can require counting a set of objects or comparing attributes across separately encountered instances. The agent must retain observations from different places and bring them together to answer.

\paragraph{EXPRESS-Bench.}
EXPRESS-Bench~\cite{fine_eqa} pairs open-ended questions with reference answers and exploration trajectories in Habitat scenes. The trajectories record actions and observations that lead to question-relevant evidence. The benchmark's Exploration-Answer Consistency measure considers whether the answer is supported by evidence from the explored scene as well as whether it agrees with the reference.

\paragraph{R2R-CE.}
Room-to-Room (R2R)~\cite{r2r} pairs indoor routes with human-written instructions that refer to visible landmarks, turns, and room transitions. R2R-CE~\cite{vlnce} converts reference routes into traversable paths in continuous Habitat scenes. The agent follows the instruction through low-level motion and receives views determined by its executed path.

\paragraph{RxR-CE.}
Room-Across-Room (RxR)~\cite{rxr} provides route instructions in English, Hindi, and Telugu. Human demonstrations align words with camera poses along each route, linking language to the landmarks and viewpoints encountered during navigation. RxR-CE adapts these routes to continuous Habitat environments, where the agent executes local movements to follow the instructions.

\subsection{Baselines}
\label{app:baselines}

\subsubsection{Interaction Tasks}
\label{app:baselines_interaction}
The baselines in \cref{tab:spatialworld} use Gemini-3.1-Pro, Qwen-3.5-397B-A17B, Qwen3.8-27B, GPT-5, and GPT-6.1-Sol as decision policies under the native SpatialWorld harness, which passes the task instruction, an egocentric RGB image, and recent interaction history to the policy~\cite{spatialworld}. The policy issues symbolic commands for movement, viewpoint changes, object interaction, or completion. The harness maps those commands to simulator-specific operations. 

\Cref{tab:embodiedeval} reports the original benchmark's native-harness results for GPT-4o and Gemini-Pro-1.5~\cite{embodiedeval} alongside results from the ground-truth-reward variant of SEEA-R1~\cite{seea_r1}. EmbodiedEval's native harness presents an egocentric image history and execution feedback~\cite{embodiedeval}. The policy selects a numbered movement, interaction, or answer option. Movement follows a scene-specific navigation graph, and available interactions and answers depend on the current task.

\begin{itemize}
    \item \textbf{SEEA-R1}~\cite{seea_r1} fine-tunes Qwen2.5-VL-7B-Instruct with trajectories collected by Monte Carlo tree search. Tree-GRPO updates the policy by comparing action values within each search node. Its ground-truth-reward variant trains on simulator rewards from ALFWorld and executes a single action path when transferred to EmbodiedEval.
\end{itemize}

\subsubsection{Vision-and-Language Navigation}
\label{app:baselines_vln}
\paragraph{Training-based methods.}
\begin{itemize}
    \item \textbf{NaVid}~\cite{navid} encodes the observation stream as spatiotemporal context and predicts the next navigation action from the video history and instruction.
    \item \textbf{Uni-NaVid}~\cite{uninavid} learns a shared video-based policy across navigation tasks. It merges redundant visual tokens online across space and time before predicting low-level actions.
    \item \textbf{NaVILA}~\cite{navila} generates mid-level commands in natural language with distances or turn angles, then executes them using a visual locomotion policy trained with reinforcement learning.
    \item \textbf{JanusVLN (Extra)}~\cite{janusvln} caches visual-semantic and spatial-geometric features separately for navigation. The Extra variant trains with ScaleVLN and DAgger data in addition to R2R-CE and RxR-CE.
    \item \textbf{StreamVLN}~\cite{streamvln} retains recent interaction turns in a sliding window and compresses older visual context through 3D-aware token pruning.
    \item \textbf{DualVLN}~\cite{dualvln} grounds intermediate goals in image pixels with a VLM planner. A faster diffusion policy uses those goals and features from the planner to generate motion trajectories.
    \item \textbf{InternVLA-N1}~\cite{internvlan1} learns latent plans after pretraining on pixel goals. Its diffusion policy receives these plans asynchronously to generate short-horizon trajectories.
    \item \textbf{NavFoM}~\cite{navfom} identifies cameras and time steps with tokens and dynamically samples observations, allowing one navigation policy to serve multiple tasks and robot embodiments.
    \item \textbf{ABot-N0}~\cite{abotn0} pairs task-conditioned reasoning in an LLM with a flow-matching action head that samples continuous waypoints and headings.
    \item \textbf{OmniNav}~\cite{omninav} uses its fast visual policy for VLN. The policy predicts continuous waypoints with a flow-matching head.
    \item \textbf{AstraNav-World}~\cite{astranavworld} models future observations with action-conditioned video diffusion and predicts navigation trajectories conditioned on the generated visuals.
    \item \textbf{Qwen-RobotNav}~\cite{qwenrobotnav} use a Qwen3-VL backbone with a navigation action head that predicts waypoint trajectories. The system also supports an upper-level planner that selects the navigation mode and controls how visual history is encoded.
\end{itemize}

\paragraph{Zero-shot agents.}
\begin{itemize}
    \item \textbf{Open-Nav}~\cite{opennav} breaks instructions into actions and landmarks and tracks progress through its navigation history. It uses descriptions of objects and spatial relations to choose among predicted navigable waypoints.
    \item \textbf{Three-Step Nav}~\cite{threestepnav} plans from instruction landmarks, aligns the current observation with the next subgoal, and reviews the completed trajectory for drift before stopping.
    \item \textbf{EvoNav}~\cite{evonav} predicts upcoming actions and landmarks to estimate progress and retrieves experience from earlier trajectories and scene observations to guide the next move.
    \item \textbf{SmartWay}~\cite{smartway} pairs a waypoint predictor trained with occupancy supervision with a navigator that uses its history to revise the plan and backtrack when necessary.
    \item \textbf{AgenticNav}~\cite{agenticnav} gives a VLM tools to navigate toward pixel-selected destinations, query depth at selected pixels, and retrieve earlier observations using a compact trajectory map.
    \item \textbf{HarnessVLN}~\cite{harnessvln} checks proposed actions for evidential support, geometric feasibility, and consistency with the current subgoal. An event hierarchy and a spatiotemporal graph retain progress, spatial evidence, and execution failures.
    \item \textbf{Minimal agent}~\cite{agents_control} gives a general-purpose model separate tools to request an observation and to execute a sequence of primitive forward, turn, or stop actions.
    \item \textbf{GPT-6-Astra workflow}~\cite{astra_vln} assigns task planning, progress assessment, action proposal, and arrival assessment to separate model calls. The workflow retains observations and progress records and validates proposed actions before execution.
\end{itemize}

\subsubsection{Object-Goal Navigation}
\label{app:baselines_objectnav}
\begin{itemize}
    \item \textbf{VLFM}~\cite{vlfm} identifies exploration frontiers on an occupancy map and ranks them with a spatial value map built from image--text similarity scores.
    \item \textbf{OpenFMNav}~\cite{openfmnav} extracts candidate object concepts from free-form instructions, detects instances with vision--language foundation models, and uses a semantic score map to guide exploration.
    \item \textbf{SG-Nav}~\cite{sgnav} reasons over an online scene graph of objects, groups, and rooms. It aggregates evidence across observations to verify potential targets.
    \item \textbf{TriHelper}~\cite{trihelper} uses three specialized helpers to reroute around collisions, resume exploration when frontier selection stalls, and verify candidate detections with a VLM.
    \item \textbf{WMNav}~\cite{wmnav} projects VLM predictions of target likelihood into a persistent Curiosity Value Map. It uses directional scores and feedback on intermediate subtasks to select motion candidates.
    \item \textbf{CogNav}~\cite{cognav} maintains a heterogeneous cognitive map and prompts an LLM to choose among five search and verification states, each with its own navigation strategy.
    \item \textbf{STEGNav}~\cite{stegnav} stores object hypotheses and reachable frontiers in a spatio-temporal event graph. Recent trajectories and verified outcomes inform subsequent goal selection.
\end{itemize}
Uni-NaVid and Qwen-RobotNav-8B are described in \cref{app:baselines_vln}. The EmbodiedEval agents are described in \cref{app:baselines_interaction}.

\subsubsection{Lifelong Tasks}
\label{app:baselines_lifelong}
\begin{itemize}
    \item \textbf{Modular GOAT}~\cite{goatbench} projects semantic detections onto a top-down map and stores egocentric views and CLIP features for individual objects. It matches image goals by keypoint correspondence and text descriptions by CLIP similarity, then navigates to the matched instance.
    \item \textbf{VLMnav}~\cite{vlmnav} derives feasible motion candidates from depth and uses a map of explored space to favor new areas. It draws the candidates on the POV image for a frozen VLM to select an action.
    \item \textbf{DyNaVLM}~\cite{dynavlm} samples reachable viewpoints from RGB-D and uses a VLM to refine them. Its policy retrieves object locations and spatial relations from a graph memory updated during navigation.
    \item \textbf{TANGO}~\cite{tango} turns instructions into executable programs over perception and navigation primitives. For each new goal, it scores stored visual features against the goal embedding to recompute the exploration value map.
    \item \textbf{MTU3D}~\cite{mtu3d} merges object queries from RGB-D into spatial memory and extracts frontiers from an occupancy map. A trained spatial reasoning transformer scores remembered objects and frontiers together to select a destination.
    \item \textbf{3D-Mem}~\cite{3dmem} groups co-visible objects into image snapshots and represents unexplored directions with frontier snapshots. It filters stored views by goal-relevant categories so a VLM can reason over retained evidence and choose where to explore next.
    \item \textbf{EvoMemNav}~\cite{evomemnav} organizes images and object hypotheses in a room--view--object graph. It shortlists viewpoints for VLM verification and uses subtask outcomes to update goal-conditioned priors on the graph.
    \item \textbf{MSGNav}~\cite{msgnav} uses images to encode relations in a 3D scene graph and retrieves a subgraph relevant to the goal. Once it identifies a target, it checks occlusion against the scene point cloud to choose a final viewpoint.
\end{itemize}

\subsubsection{Embodied Question Answering}
\label{app:baselines_eqa}
\begin{itemize}
    \item \textbf{Explore-EQA}~\cite{exploreeqa} projects VLM scores for visually prompted directions onto a semantic map built from depth. It uses these scores to sample exploration frontiers and applies multi-step conformal prediction to decide when the evidence supports a single answer.
    \item \textbf{SayPlanVision}~\cite{grapheqa} adapts SayPlan for question answering by planning over a complete 3D scene graph built offline. It uses task-relevant images and stops when answer confidence reaches a threshold.
    \item \textbf{GraphEQA}~\cite{grapheqa} updates a hierarchical 3D metric-semantic scene graph online and retains images relevant to the question. A VLM uses both sources to assess answer confidence and, if it needs more evidence, choose an object or frontier node to visit.
    \item \textbf{MemoryEQA}~\cite{memoryeqa} stores descriptions of observed entities alongside the agent's pose and indexes them with dense embeddings. Entropy-based retrieval adjusts the amount of scene evidence supplied to the planning, stopping, and answering modules.
    \item \textbf{Fine-EQA}~\cite{fine_eqa} combines a global semantic map with a map of functional regions. The agent explores frontiers until it identifies a region relevant to the question, then searches within that region using a masked semantic map.
    \item \textbf{FAST-EQA}~\cite{fasteqa} prioritizes openings and doorways during global exploration and searches locally for targets inferred from the question. It retains a bounded set of relevant images per target for answer reasoning.
\end{itemize}
The EmbodiedEval agents, Qwen-RobotNav-8B, and 3D-Mem are described in \cref{app:baselines_interaction,app:baselines_vln,app:baselines_lifelong}.

\subsection{Computing Infrastructure}
\label{app:computing}
Our inference infrastructure includes a node equipped with eight PPU-ZW805 accelerators, each with 96\,GiB of device memory. We serve Qwen3.8-27B with the PPU-adapted vLLM release \texttt{0.18.0+ppu2.0.0.oe}, using four accelerators per instance with four-way tensor parallelism. Closed-model evaluations use hosted APIs.
Simulation runs on a dual-socket AMD EPYC 7763 host with 128 physical CPU cores and approximately 504\,GiB of usable RAM.

\subsection{Experiment Setup}
\label{sec:settings}

\subsubsection{Evaluation configuration}
\label{sec:settings-observations}

\Cref{tab:habitat_eval_config} gives the POV camera parameters and rollout limits for the Habitat evaluations.

\begin{table}[t]
\centering
\captionsetup{singlelinecheck=false}
\caption{\textbf{Habitat POV cameras and rollout budgets.} POV is $640\times480$ and height is relative to the agent base. Budgets count policy decisions and low-level environment actions per episode or question. GOAT allows 30 decisions for the first subtask and 20 for each later one, with a shared episode action budget.}
\label{tab:habitat_eval_config}
\setlength{\tabcolsep}{2pt}
\begin{tabularx}{\linewidth}{@{}L{0.34\linewidth}*{2}{>{\centering\arraybackslash}p{0.11\linewidth}}*{2}{>{\centering\arraybackslash}X}@{}}
\toprule
\multirow{2}{*}{\textbf{Benchmark}} & \multicolumn{2}{c}{\textbf{POV camera}} & \multicolumn{2}{c}{\textbf{Rollout budget}} \\
\cmidrule(lr){2-3}\cmidrule(lr){4-5}
& HFOV & Ht. (m) & Decisions & Actions \\
\midrule
R2R-CE / RxR-CE & $110^\circ$ & 1.31 & 30 & 500 \\
HM3D-v2 & $120^\circ$ & 1.50 & 30 & 500 \\
MP3D & $79^\circ$ & 0.88 & 60 & 500 \\
GOAT-Bench & $110^\circ$ & 1.31 & 30 / 20 & 5,000 \\
\mbox{HM-EQA / MT-HM3D} & $120^\circ$ & 1.50 & 30 & 1,000 \\
EXPRESS-Bench & $120^\circ$ & 1.50 & 30 & 500 \\
\bottomrule
\end{tabularx}
\end{table}

\subsubsection{Navigation scoring}
\label{sec:settings-vln}

R2R-CE uses the Open-Nav 100-episode subset~\cite{opennav,agenticnav,harnessvln}. RxR-CE uses 100 randomly sampled English guide tasks from val-unseen.

\paragraph{Object-goal navigation and lifelong tasks.}
\label{sec:settings-objectnav}
\label{sec:settings-lifelong}
HM3D-v2 and MP3D evaluate explicit stops across 1,000 and 2,195 tasks, respectively. GOAT-Bench evaluates each \texttt{subtask\_stop}. These evaluations measure geodesic distance to the nearest released target viewpoint. We aggregate SR and SPL per episode for ObjectNav and over subtasks for GOAT-Bench.

\paragraph{Shared metrics.}
For $N$ tasks or subtasks, let $S_i$ indicate success and let $\ell_i^\star$ and $\ell_i$ denote the shortest valid path and the executed path length. The navigation metrics are
\begin{align}
 \mathrm{SR}&=\frac{100}{N}\sum_{i=1}^{N}S_i,
 \label{eq:sr}\\
 \mathrm{SPL}&=\frac{100}{N}\sum_{i=1}^{N}S_i\frac{\ell_i^\star}{\max(\ell_i^\star,\ell_i)}.
 \label{eq:spl}
\end{align}
For $\ell_i^\star=0$, the per-task SPL contribution is $S_i$.

\paragraph{VLN trajectory metrics.}
NE and OSR use the geodesic distance $d(p_{it},G_i)$ from a recorded position to the goal set. OSR includes the start and the reached positions. For nDTW and sDTW, $D_i$ is the FastDTW cost with radius 1 and 3D Euclidean point distances between the deduplicated executed route and the reference route $R_i$. Here $\tau$ is the evaluation radius, and $q_i=0$ if either route is empty.
\begin{align}
 \mathrm{NE}&=\frac{1}{N}\sum_{i=1}^{N}d(p_{iT_i},G_i),
 \label{eq:settings-ne}\\
 \mathrm{OSR}&=\frac{100}{N}\sum_{i=1}^{N}\mathbf{1}\!\left[\min_t d(p_{it},G_i)<\tau\right].
 \label{eq:settings-osr}
\end{align}
\begin{align}
 q_i&=\exp\!\left(-\frac{D_i}{|R_i|\tau}\right),
 \label{eq:settings-ndtw-episode}\\
 \mathrm{nDTW}&=\frac{100}{N}\sum_{i=1}^{N}q_i,
 \label{eq:settings-ndtw}\\
 \mathrm{sDTW}&=\frac{100}{N}\sum_{i=1}^{N}S_iq_i.
 \label{eq:settings-sdtw}
\end{align}

\subsubsection{Embodied question answering}
\label{sec:settings-eqa}

HM-EQA and MT-HM3D use exact option-label accuracy $100\sum_i\mathbf{1}[\hat y_i=y_i]/N$ over all assigned questions, with missing answers counted as incorrect. EXPRESS uses free-form answers. A separate Qwen3.8-Flash-Next judge reads the question, reference answer, prediction, and final POV, and assigns a visual-grounding rating $g_i\in\{0,0.5,1\}$ and an answer-quality rating $a_i\in\{1,\ldots,5\}$. Let $\mathcal{V}$ contain the scored trajectories with finite reference and executed path lengths $\ell_i^\star$ and $\ell_i$. All 2,044 assigned questions have valid judge scores and finite path lengths, so both metrics use $N=|\mathcal{V}|=2{,}044$.
\begin{align}
 C^{\star} &= \frac{100}{N}\sum_{i=1}^{N}\frac{a_i}{5},
 \label{eq:express-answer-score}\\
 E_{\mathrm{path}} &= \frac{100}{|\mathcal{V}|}\sum_{i\in\mathcal{V}}
 \frac{g_i a_i}{5}\frac{\ell_i^\star}{\max(\ell_i^\star,\ell_i)}.
 \label{eq:express-efficiency}
\end{align}

\subsubsection{Interaction and EmbodiedEval}
\label{sec:settings-interaction}

The AI2-THOR and ProcTHOR subsets use task-specific predicates to evaluate the final simulator state.

\paragraph{EmbodiedEval.}
\label{sec:settings-embodiedeval}
Overall pools all 328 tasks, including evaluator errors. With $K_i$ conditions for task $i$ and indicator $c_{ik}$ for condition $k$, goal-condition success is
\begin{equation}
 \mathrm{GcS}=\frac{100}{N}\sum_{i=1}^{N}\frac{1}{K_i}\sum_{k=1}^{K_i}c_{ik}.
 \label{eq:settings-gcs}
\end{equation}

\section{Limitations}
\label{app:limitations}

The policy parameters remain frozen throughout the whole agent loop. We leave training policies to manage memory and action interfaces to future work.

\section{Qualitative Results}
\label{app:qualitative_results}

\subsection{Successful Cases}
\label{app:successful_cases}
\Crefrange{fig:qual_success_r2r}{fig:qual_success_embodiedeval} show eleven successful rollouts across all five task families, one per benchmark or evaluated subset. Each panel opens with the task and a top-down overview. The steps below pair selected observations with actions and excerpts from the policy's decisions.

\subsection{Failure Cases}
\label{app:failure_cases}
\Crefrange{fig:qual_failure_grounding}{fig:qual_failure_environment} illustrate the five primary error types defined in \cref{sec:what_remains_difficult}, in the same order. The red annotation highlights the relevant decision. The first four cases involve failures in target selection, interaction execution, counting, and exploration. In the final case, two doorways fit the same route instruction.

\begingroup
\newcommand{\qualitativecase}[3]{%
  \begin{figure*}[p]
    \begin{minipage}[c][.99\textheight][c]{\textwidth}
      \centering
      \includegraphics[pagebox=cropbox,width=\linewidth]{figures/case_analysis/#1.pdf}
      \caption{#2}
      \label{#3}
    \end{minipage}
  \end{figure*}%
}

\qualitativecase{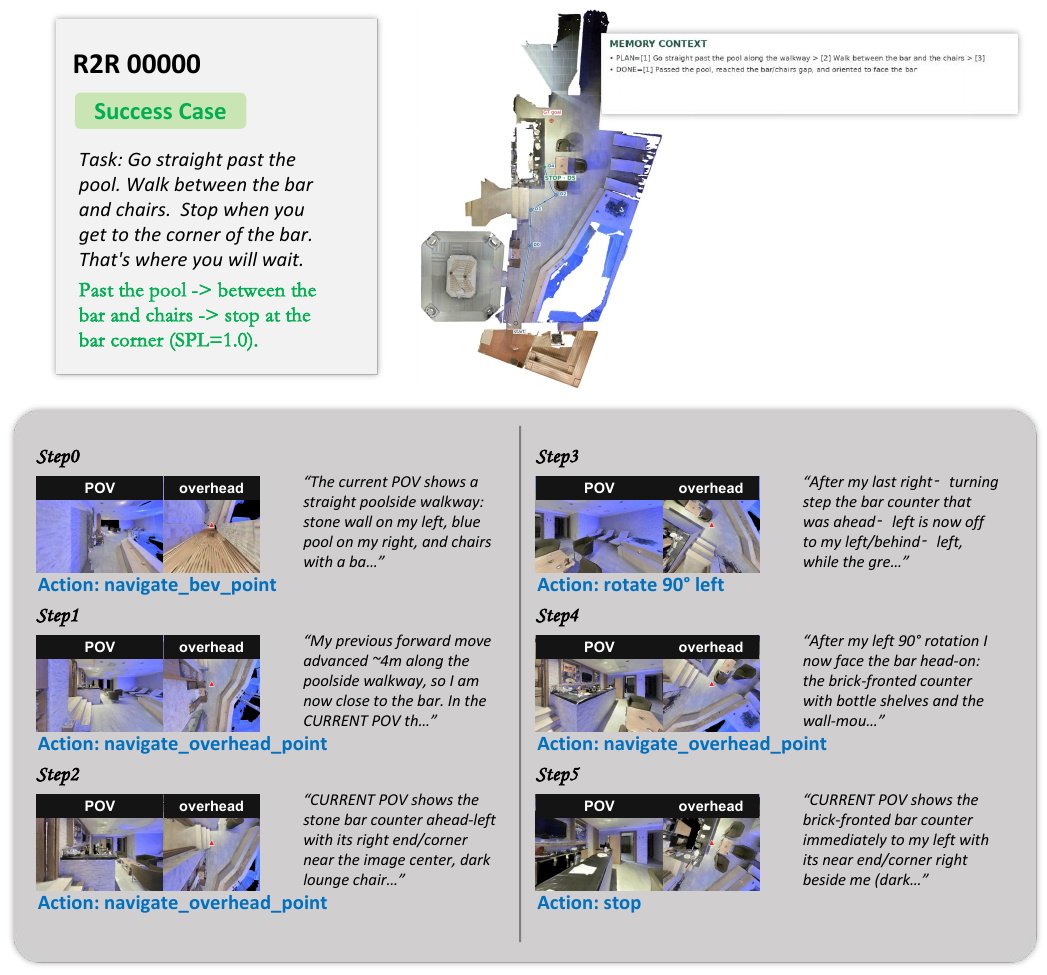}{Following the poolside route to the bar corner on R2R-CE.}{fig:qual_success_r2r}
\qualitativecase{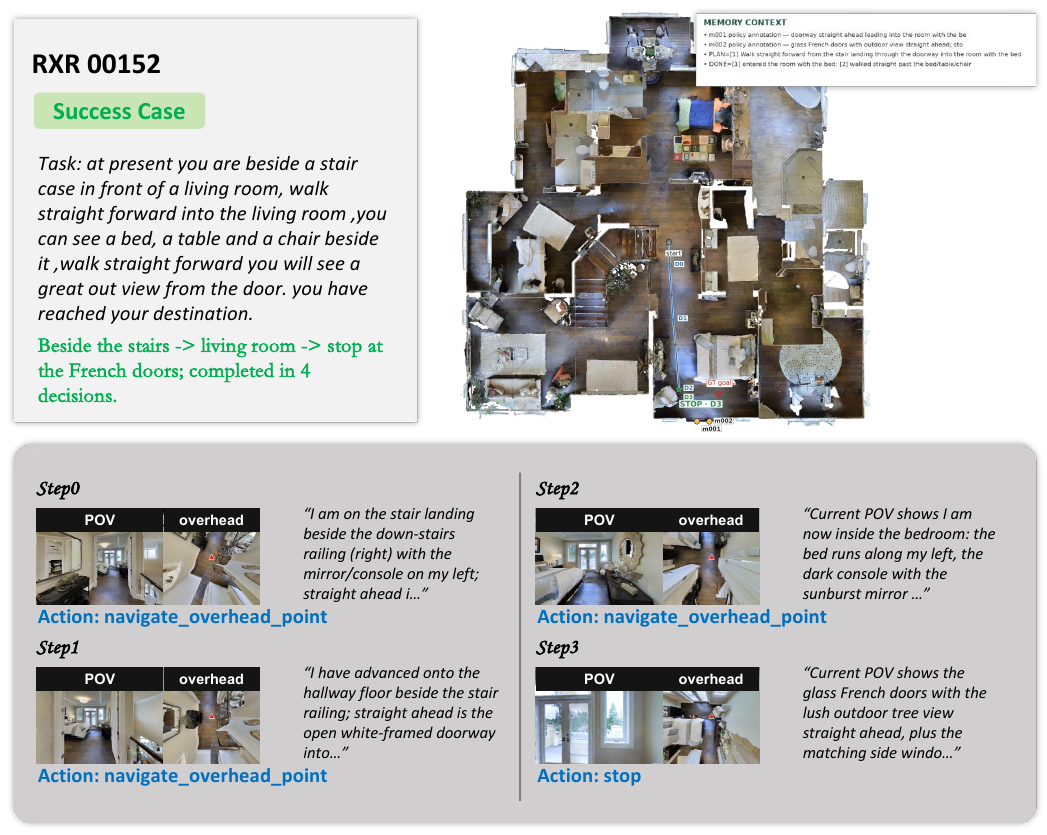}{Navigating from the stairs to the French doors on RxR-CE.}{fig:qual_success_rxr}
\qualitativecase{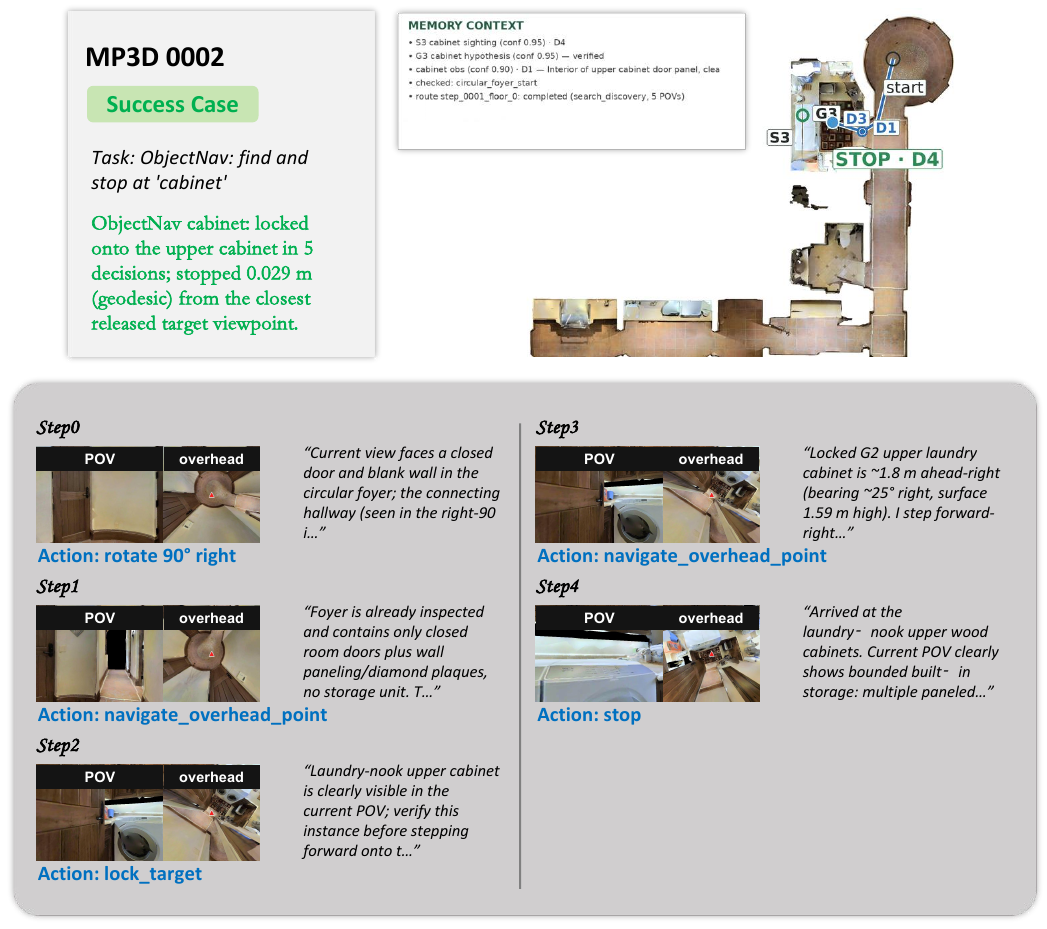}{Locating a cabinet and stopping at the target on MP3D.}{fig:qual_success_mp3d}
\qualitativecase{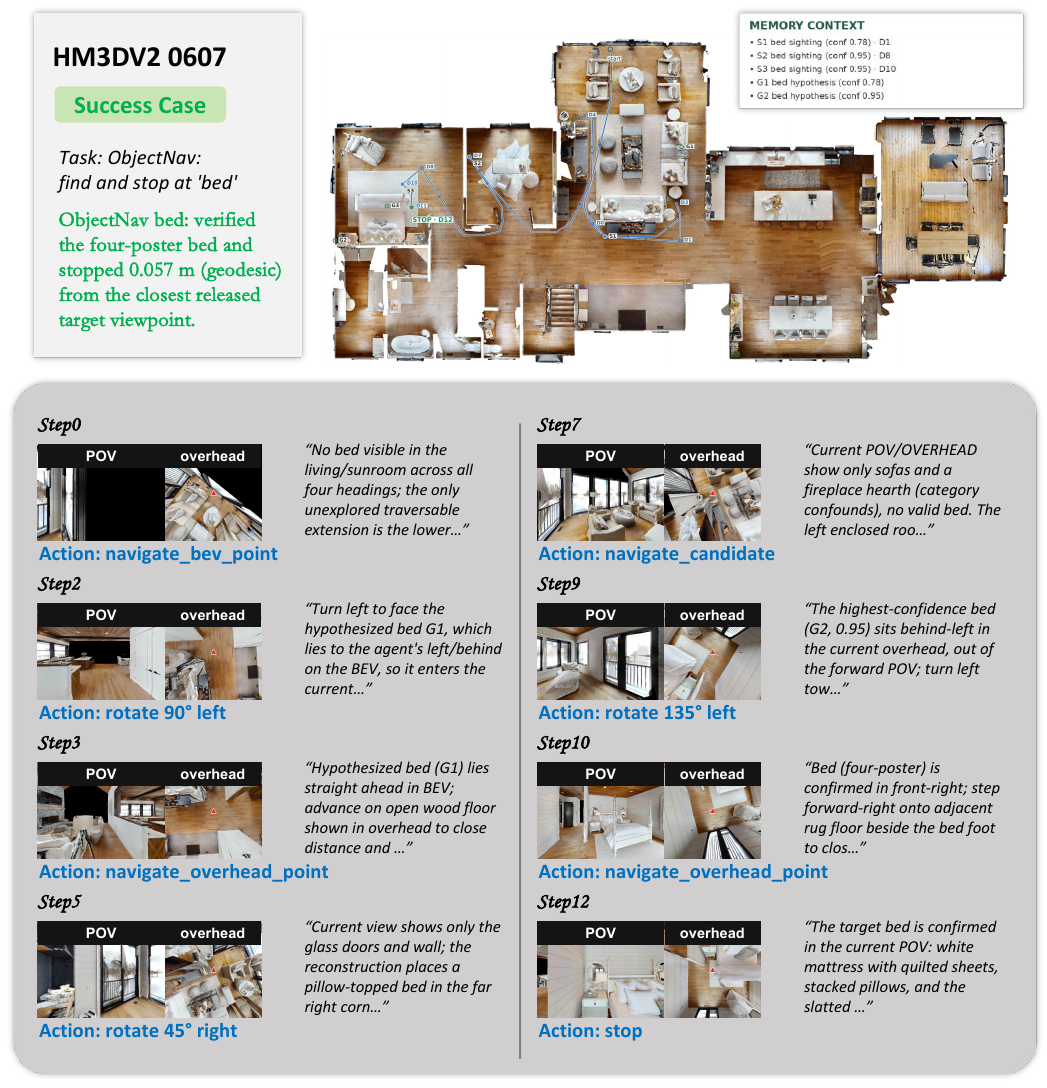}{Finding the four-poster bed on HM3D-v2.}{fig:qual_success_hm3d}
\qualitativecase{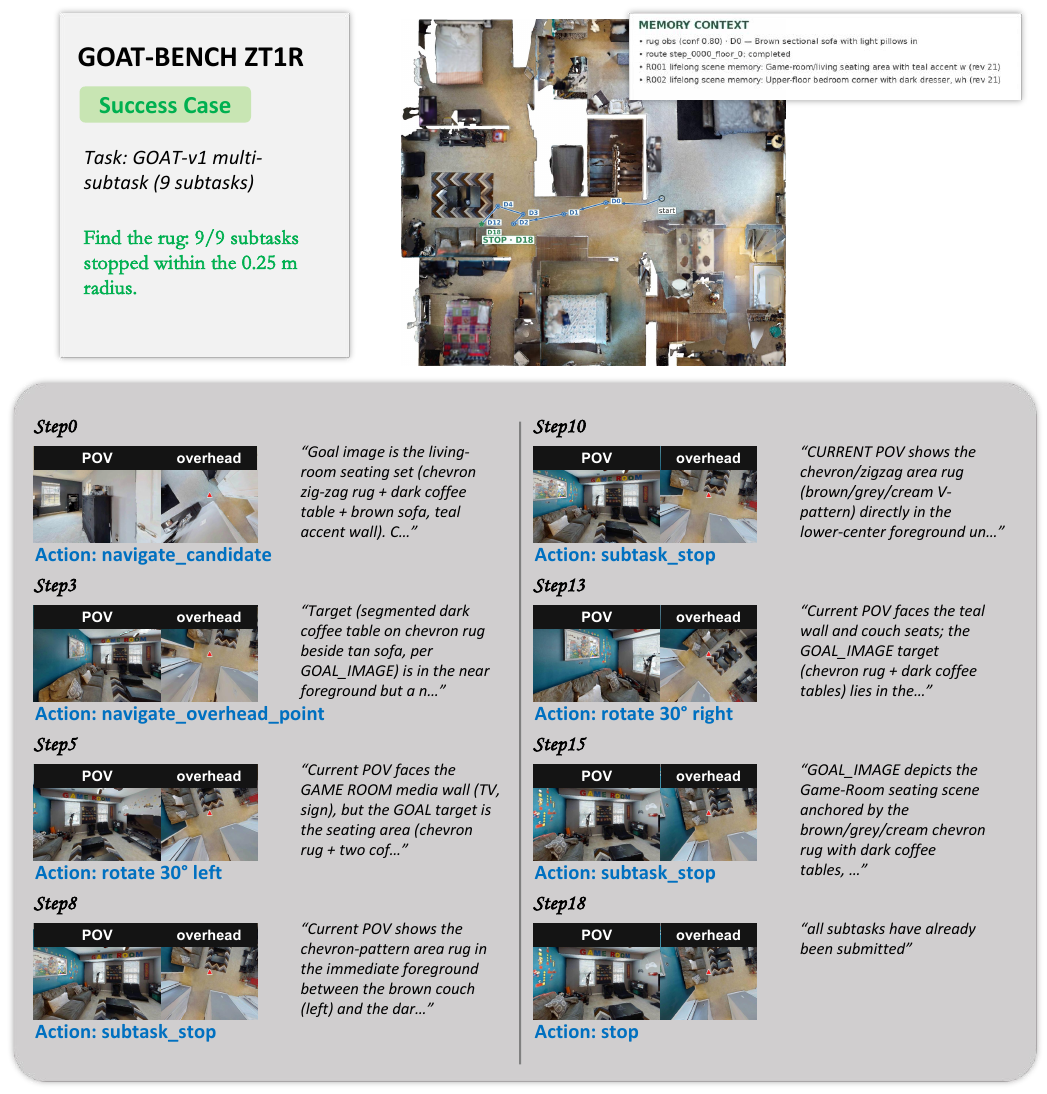}{Completing all nine consecutive goals in a GOAT-Bench episode.}{fig:qual_success_goat}
\qualitativecase{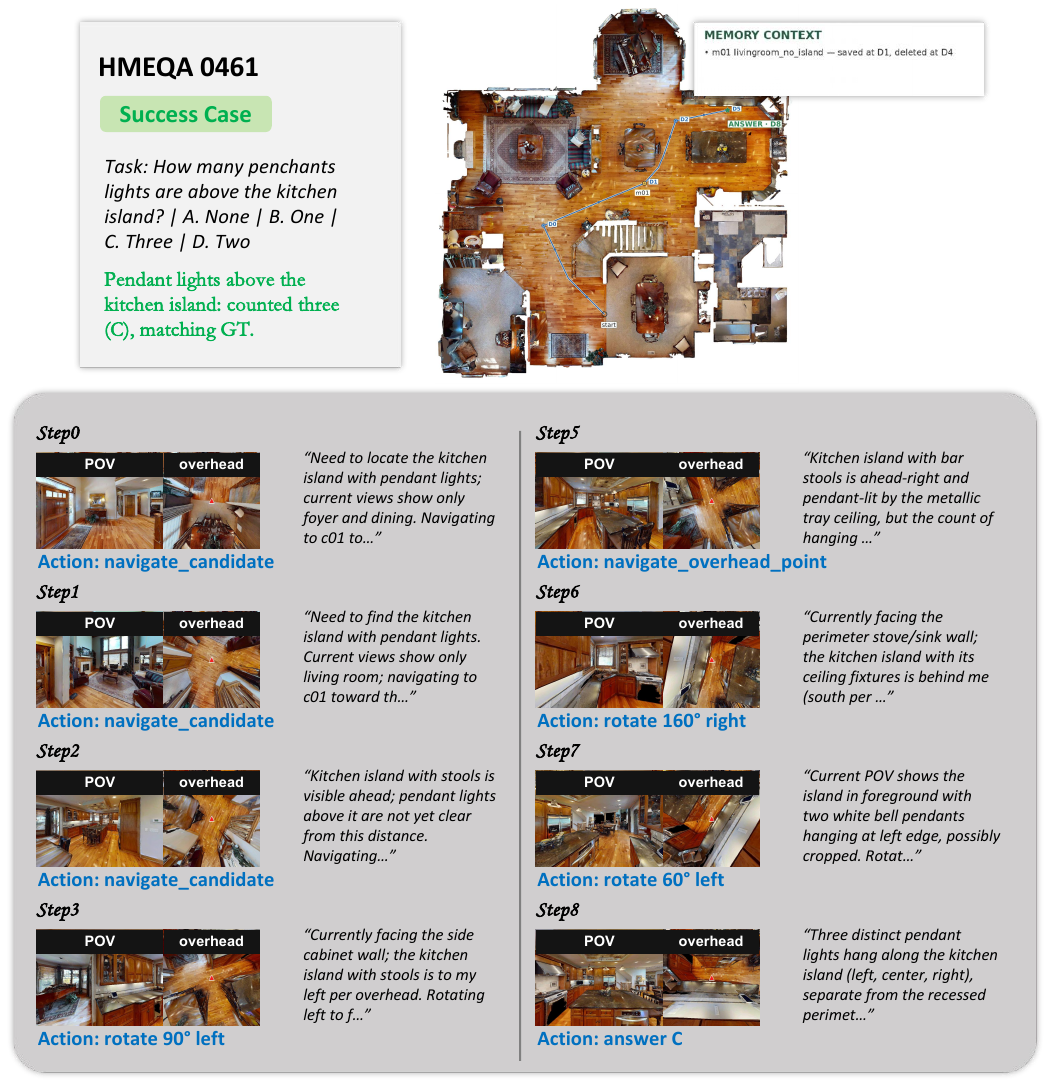}{Counting pendant lights above the kitchen island on HM-EQA.}{fig:qual_success_hmeqa}
\qualitativecase{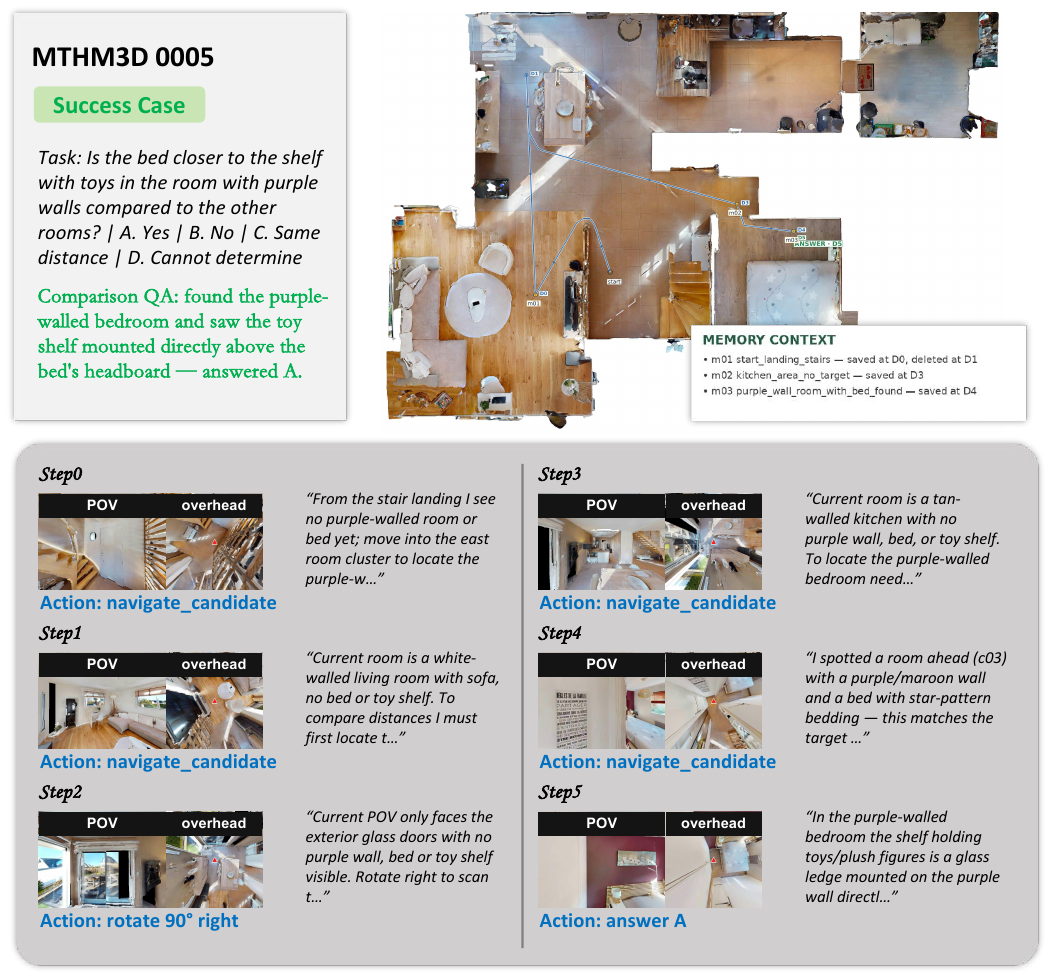}{Locating the toy shelf above the bed on MT-HM3D.}{fig:qual_success_mthm3d}
\qualitativecase{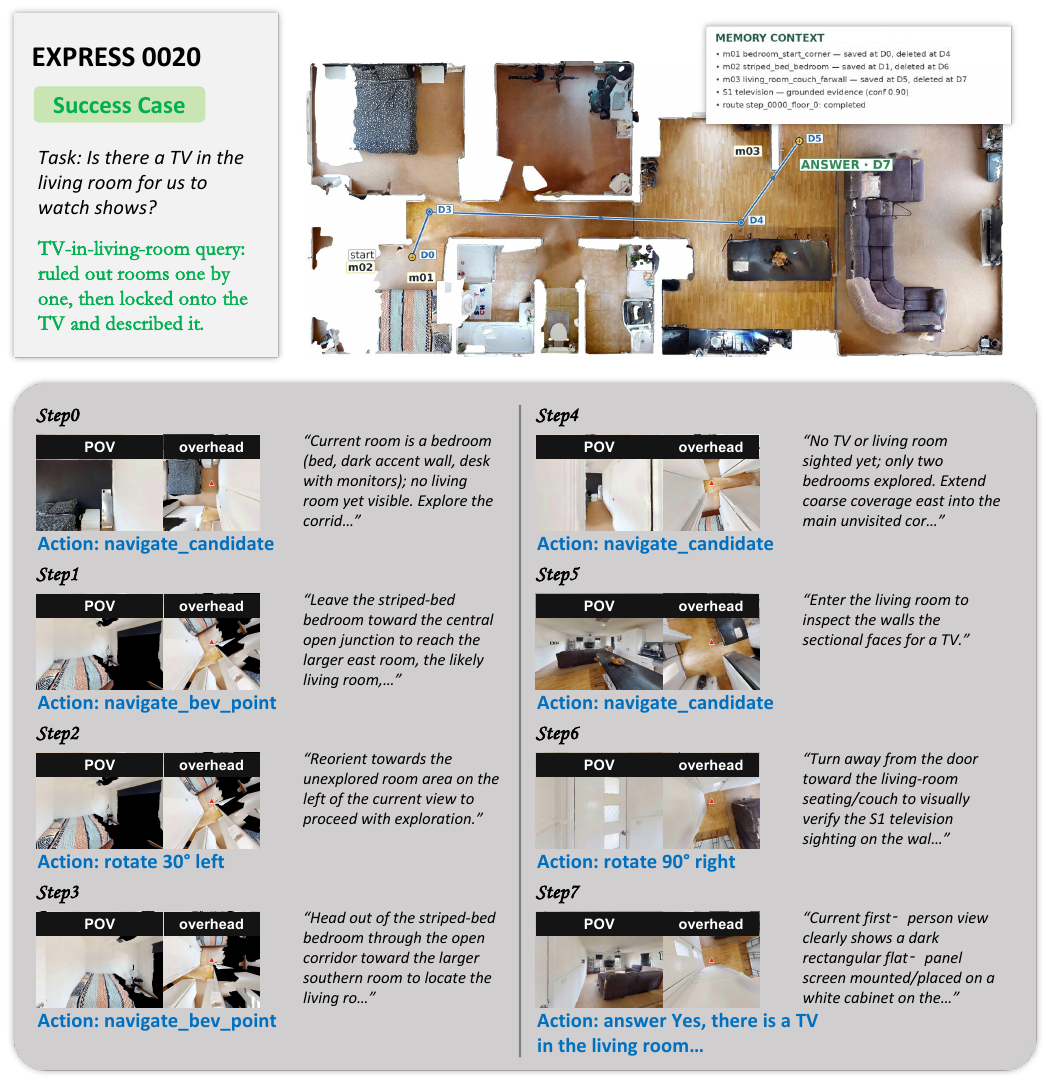}{Finding a television in the living room on EXPRESS-Bench.}{fig:qual_success_express}
\qualitativecase{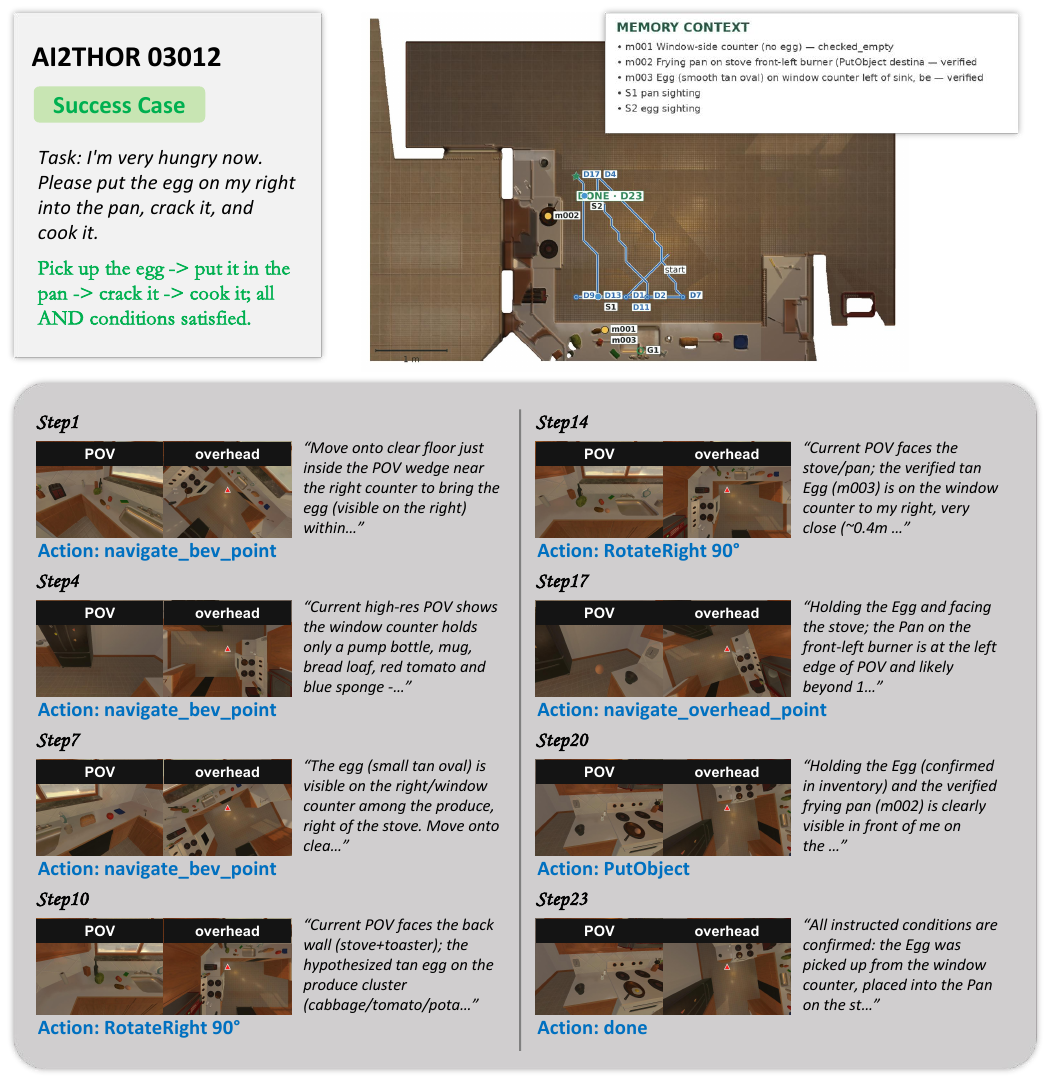}{Cooking an egg on the AI2-THOR subset of SpatialWorld.}{fig:qual_success_ai2thor}
\qualitativecase{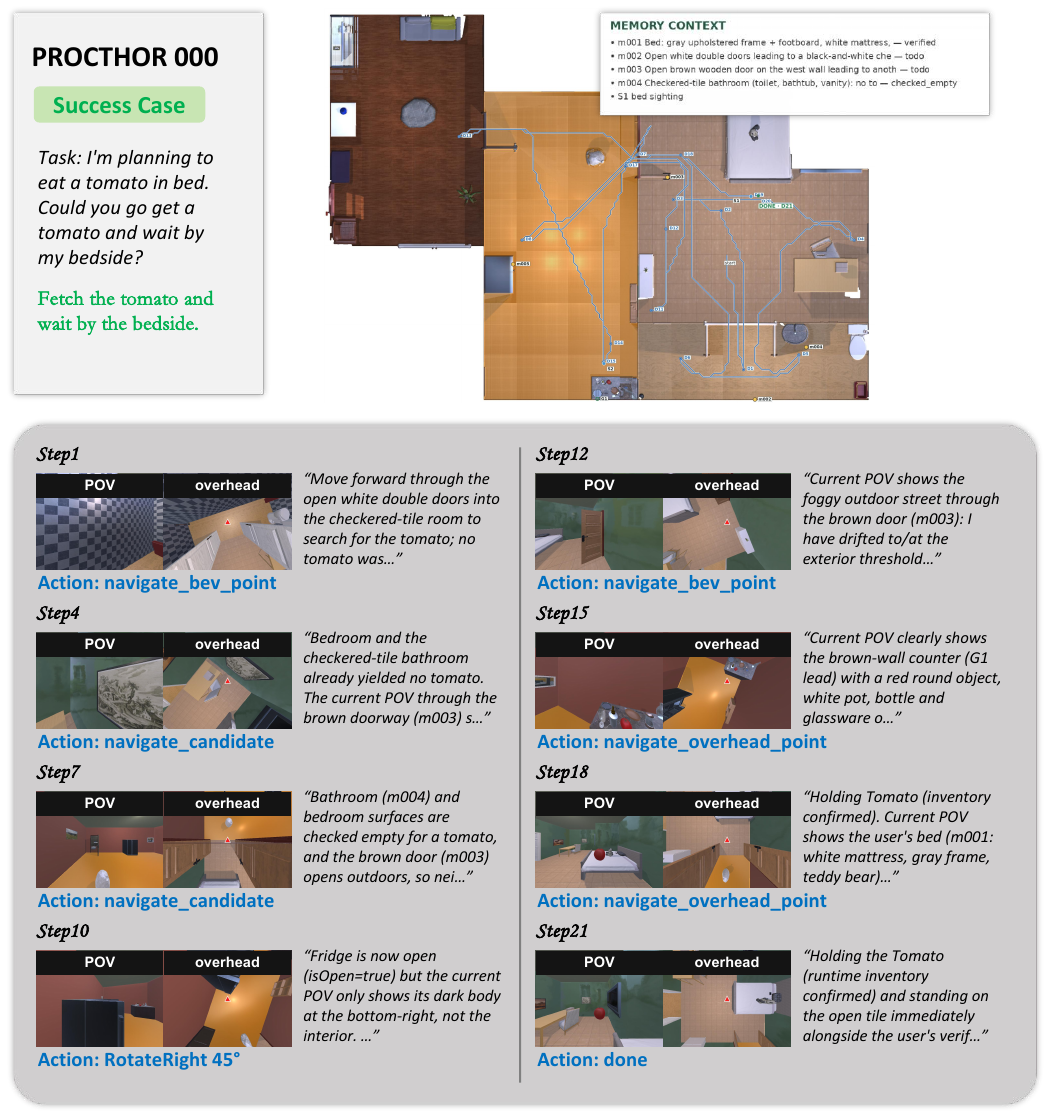}{Bringing a tomato to the bedside on the ProcTHOR subset of SpatialWorld.}{fig:qual_success_procthor}
\qualitativecase{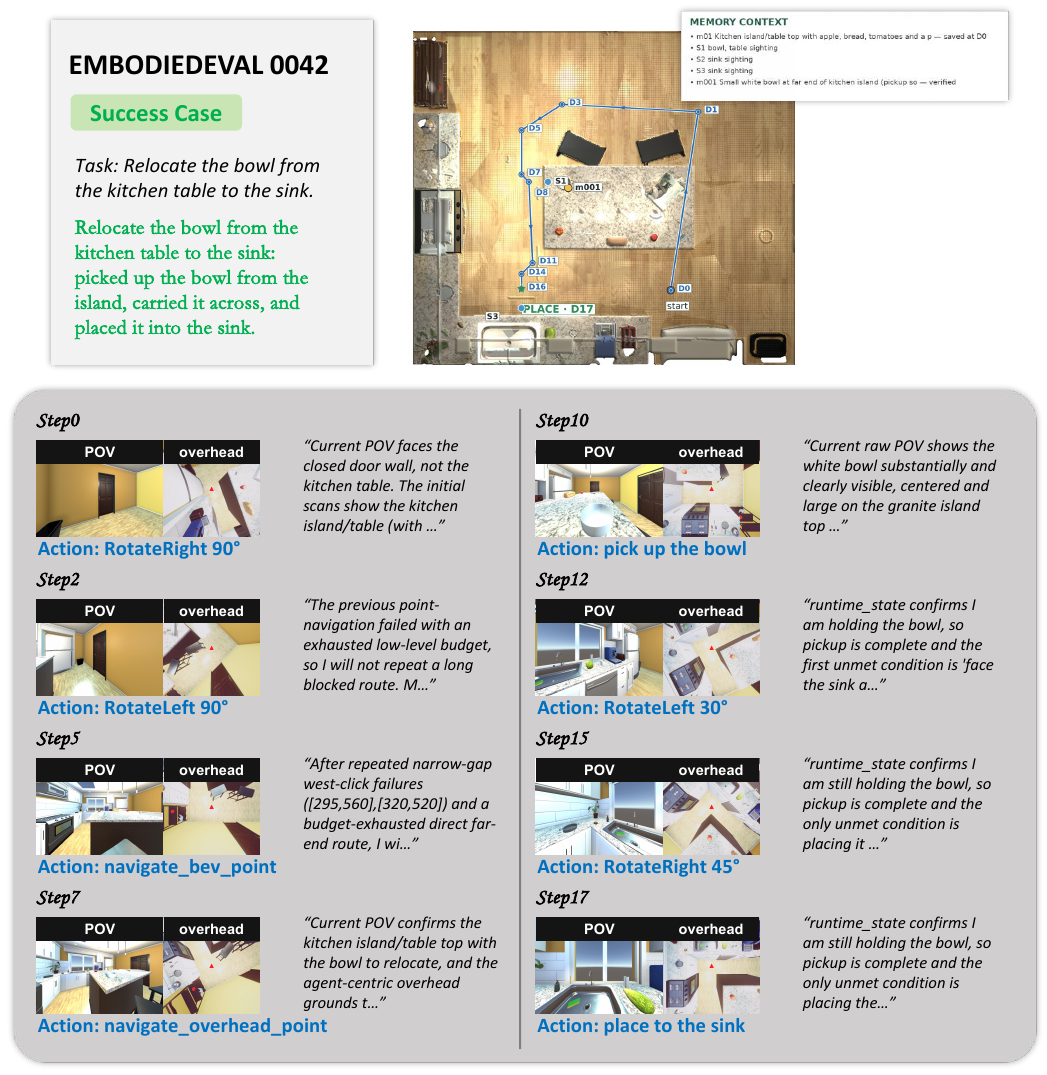}{Moving a bowl from the kitchen island to the sink on EmbodiedEval.}{fig:qual_success_embodiedeval}

\qualitativecase{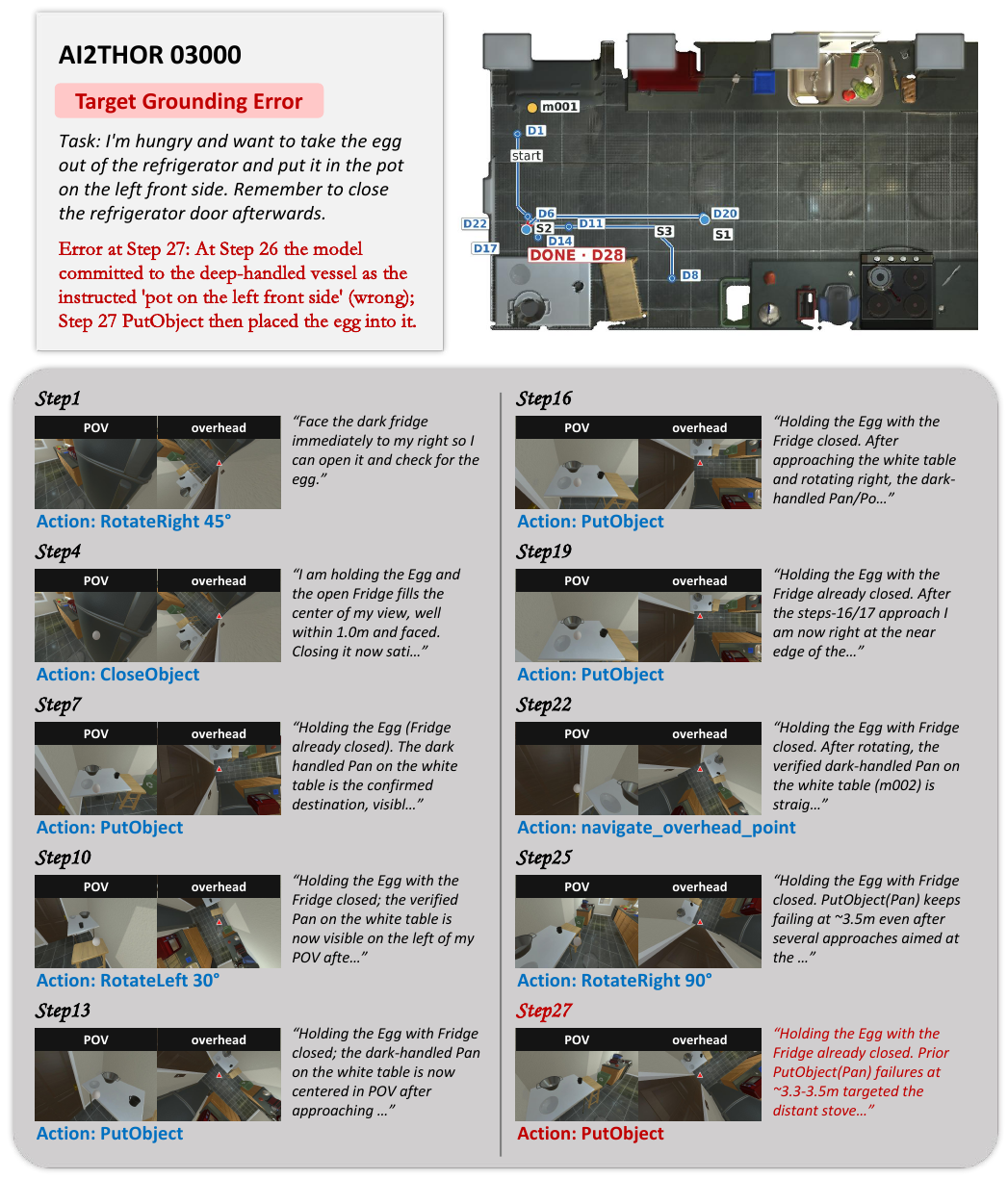}{Target grounding error on AI2-THOR when selecting the destination pot.}{fig:qual_failure_grounding}
\qualitativecase{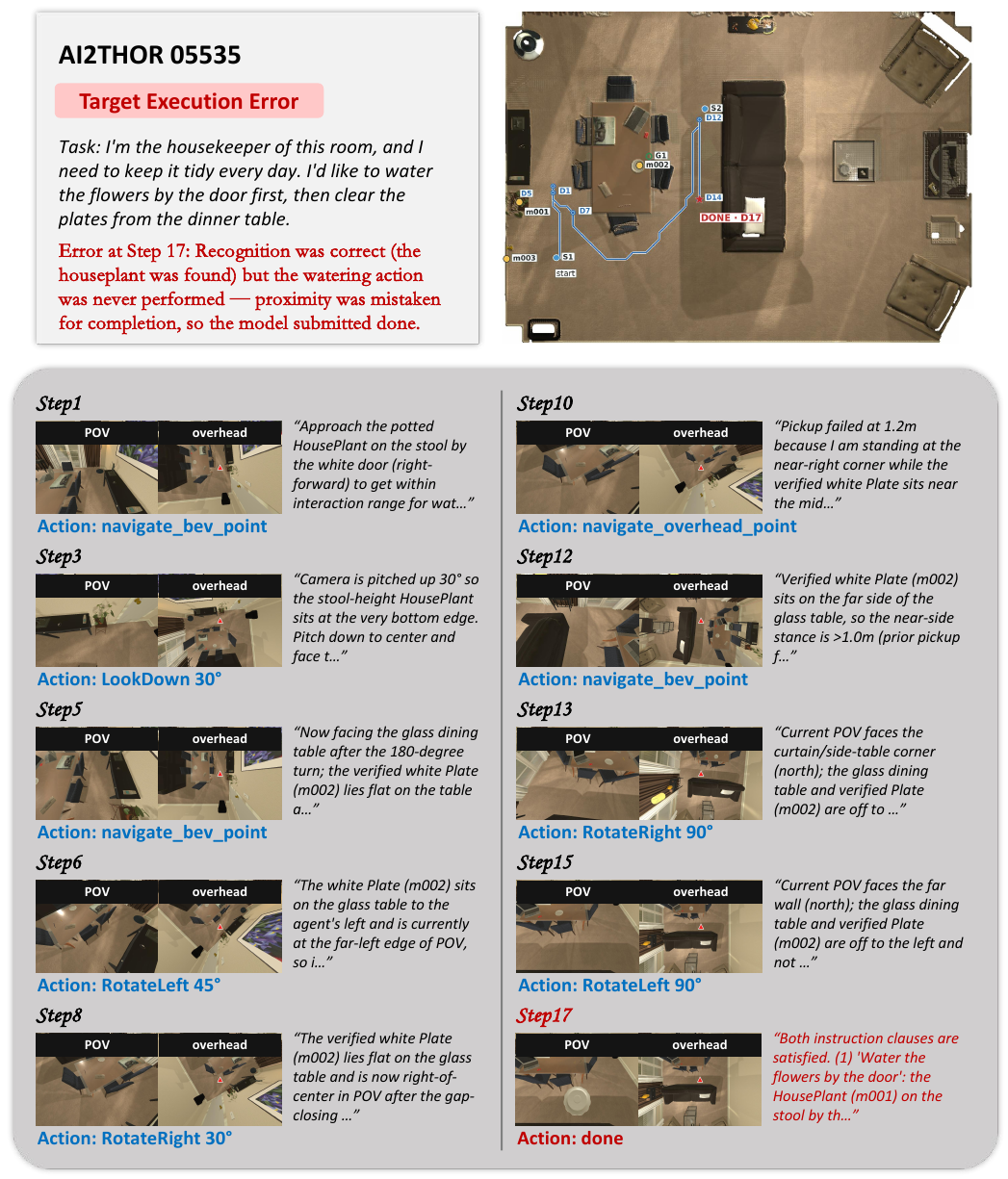}{Target execution error on AI2-THOR when the plant is approached but left unwatered.}{fig:case_a4}
\qualitativecase{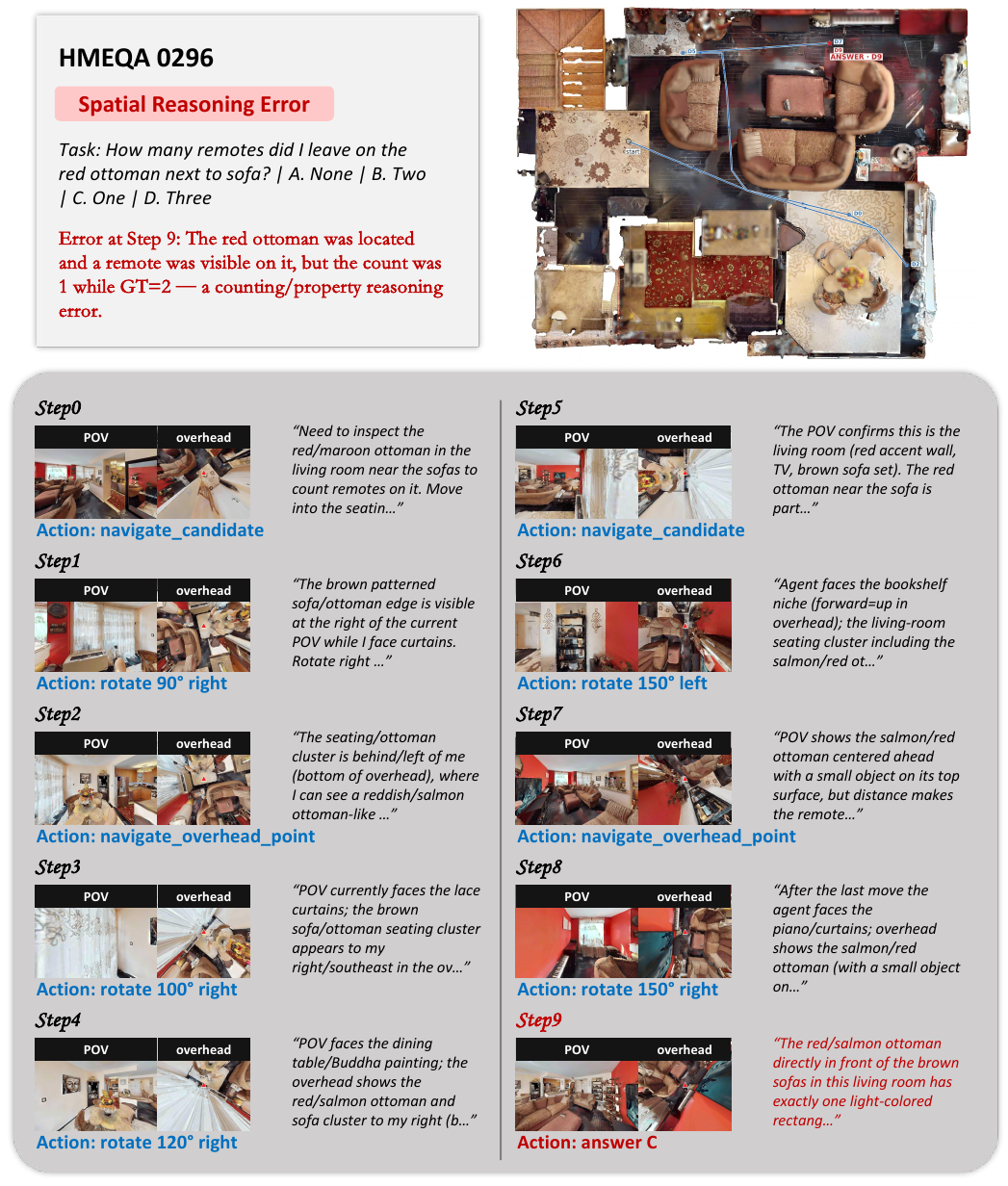}{Spatial reasoning error on HM-EQA when counting remotes on an ottoman.}{fig:qual_failure_spatial}
\qualitativecase{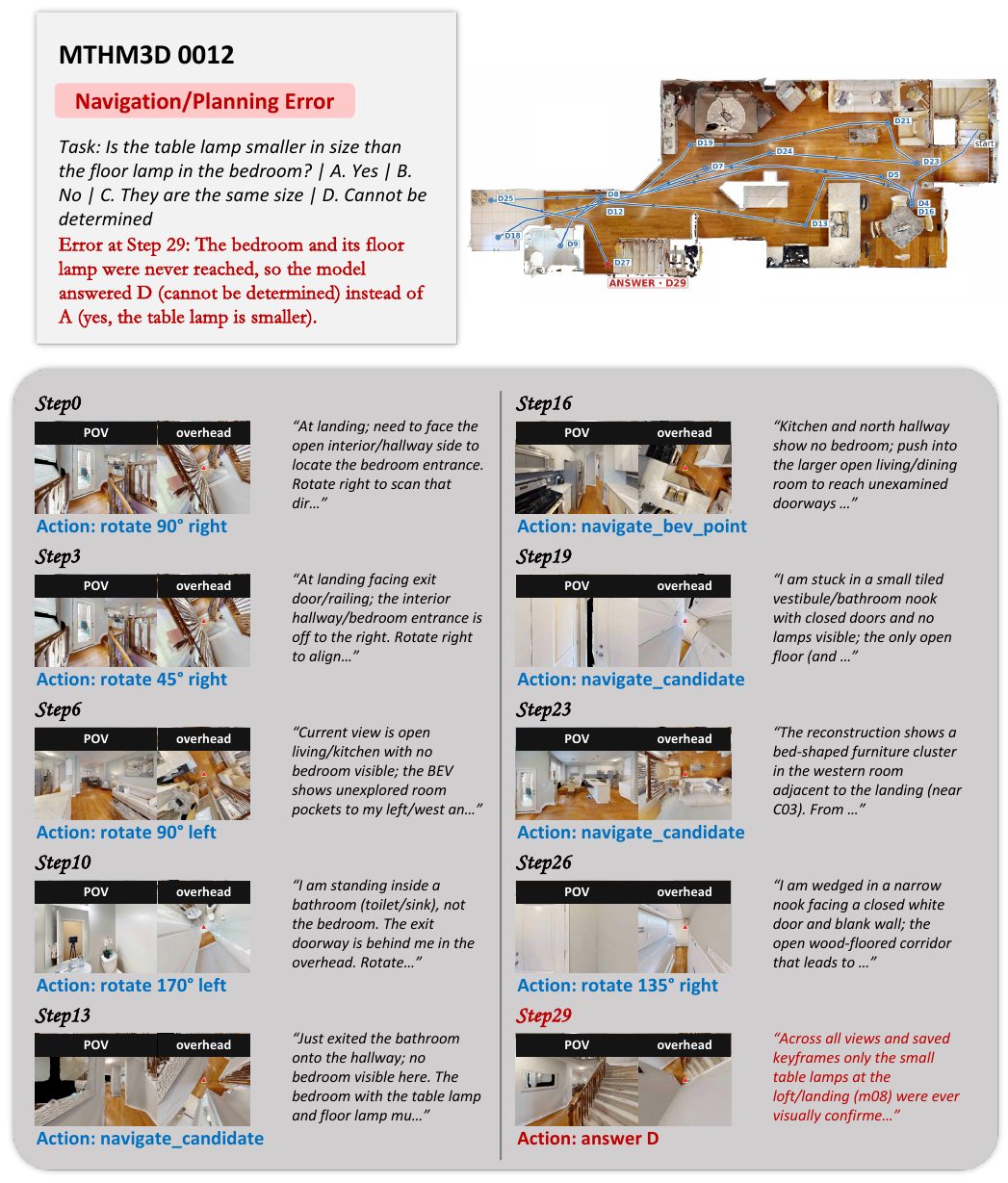}{Navigation/planning error on MT-HM3D when the agent fails to reach the bedroom.}{fig:qual_failure_planning}
\qualitativecase{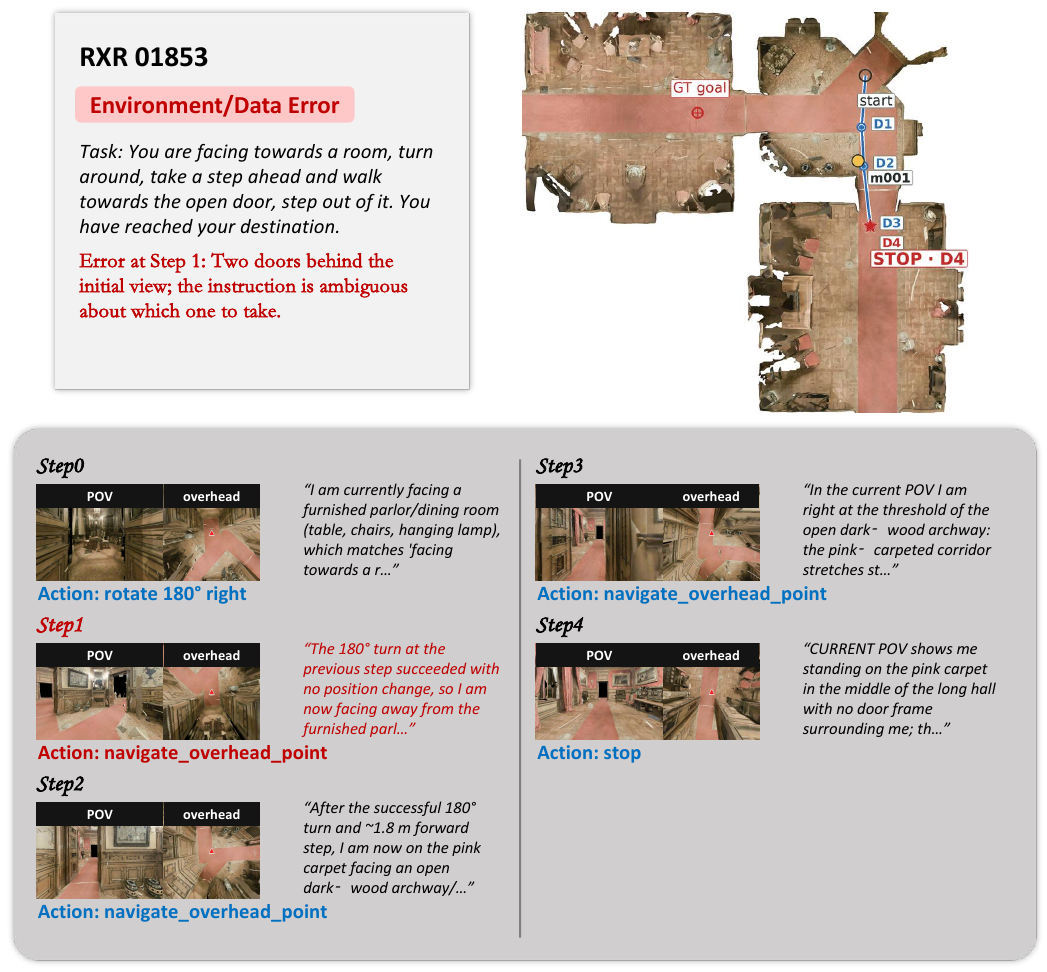}{Environment/data error on RxR-CE caused by an ambiguous doorway reference.}{fig:qual_failure_environment}
\endgroup

\onecolumn
\section{Prompts}
\label{app:prompts}

We present the prompts used for evaluation. Shared input conventions are given once below. 

\subsection{Shared inputs and conventions}
\label{app:prompt-roles}

Each decision receives the task-specific goal, ordered images (\texttt{\{IMAGE\_MANIFEST\}}), and public observations, execution feedback and memory records (\texttt{\{PUBLIC\_STATE\}}). The current action contract (\texttt{\{CURRENT\_ACTION\_CONTRACT\}}) specifies permitted actions, required fields, pointable images, valid IDs and optional memory edits.

Identify objects in current RGB views. POV shows the egocentric view. OVERHEAD shows local geometry. BEV shows occupancy and connected routes, and RECONSTRUCTION provides colored spatial context. OVERHEAD and BEV are agent-oriented, with forward at the top. Specify each point in its named current image as $[x,y]\in[0,1000]^2$ from the top-left. Match landmarks across views and refresh points after motion.

Choose observed, connected floor with clearance. Use exploration candidates for unchecked space and inspect a relevant lead when one appears. Check actual displacement and the resulting view after movement. If blocked, select a different reachable stance or intermediate point.

Use grounded evidence groups and policy annotations to guide search. Record observations available before action execution, reuse existing IDs and revise disproved hypotheses. A saved image preserves the pre-action view. Choose one permitted action in the current contract's format, with a brief reason grounded in observations and execution feedback.

A grounded evidence group links an earlier observing viewpoint to a hypothesized object location. Revisit the viewpoint to recover the viewing direction, then check the candidate object in the new image. Keep the object's location distinct from the floor position used to approach it. If an action includes a memory edit, describe evidence already available at decision time. Check the next observation for the outcome of the action and the status of the edit.

\subsection{Task-specific instructions}
\label{app:decision-prompts}

\begin{promptbox}{SpatialWorld interaction}
Complete {TASK_TEXT} in dependency order. Check the last result, inventory and first unmet condition. Approach the exact visible instance, face its accessible side and use an available interaction. Confirm pickup in inventory. Carry the item until release or placement is requested. For placement, select the destination receptacle or surface and verify the resulting state. Use done once feedback confirms all required states, holding or placement conditions, and final position. A "beside" instruction requires a stance alongside the destination.

If interaction fails, inspect the target and the reported outcome before choosing another action. Translate when the target is too distant. Change the viewing angle when a nearby target is occluded. Track attempted and completed operations separately so that a failed pickup or state change remains the next unmet condition.

AI2-THOR. Inspect the initial scans. Use room function to prioritize plausible supports and containers. Approach and open a container before examining its contents. Allow clearance for the carried object and retain evidence needed by later steps. If a useful object or support appears during exploration, inspect it before moving on. Return to a checked area when a later instruction requires it or a specific observation is still missing.

ProcTHOR. Inspect cluttered supports from closer or changed viewpoints for small targets. Verify the actual stance after a projected navigation goal. Notes derived from the same image count as one source of evidence. Mark surfaces checked after inspection and save indispensable comparison views. For a fixed fixture, identify its visible structure directly. For a small item on a cluttered surface, change distance or angle until the relevant part of the surface can be inspected. Before finishing a transport task, verify whether the instruction requires the object to remain held at the destination.
\end{promptbox}

\begin{promptbox}{EmbodiedEval}
Resolve the first unmet condition of {TASK_TEXT}. Compare the previous observation with the current image, measured motion and public feedback. A pending route leg records the destination you requested. Recheck where a partial or projected move actually ended.

Choose a stance that resolves the current dependency. To answer a spatial question, look for a view that shows the relevant relation. For a handover, approach the intended person on an accessible side. When movement changes the heading or ends short of the requested point, use the new view to locate the next approach. A view change can expose an occluded target, while excess distance requires movement.

For physical interactions, approach the exact person or object and face the interactive side. Verify pickup, placement or handover from feedback. Check whether an item is already held before attempting pickup. For choices, use the option and index displayed this turn. Follow the task's exposed completion mechanism. Track held items and failed attempts, annotate each physical object once, and frame any relation needed later before saving visual evidence.
\end{promptbox}

\begin{promptbox}{Object-goal navigation (MP3D and HM3D-v2)}
Find an instance of {GOAL_CATEGORY}. Use search candidates and remembered viewpoints to inspect unchecked areas. Identify the object in current POV from features that distinguish it from nearby lookalikes. If a feature is obscured, inspect it from another viewpoint. When available, lock_target provides current-view grounding. Use reject_target when visible evidence contradicts the proposed identity.

Use initial scans and previously observed locations to prioritize nearby rooms, openings and supports. Distinguish a room you traversed from one whose relevant surfaces you actually inspected. Give a plausible candidate a directed inspection from a useful angle. If its identity remains unresolved, retain it as a hypothesis and continue searching. Repeated annotations from one sighting do not provide independent confirmation.

Match the same object across views and choose a reachable stance on its accessible side. A partition, support edge or change of level may separate the agent from a visible target. Trace connected floor around the barrier, then check actual displacement and the resulting view. If the command succeeds with zero movement, choose another reachable stance. When current views confirm the target and show that you are beside it, issue an explicit stop. Correct any disproved target record using the available memory actions.
\end{promptbox}

\begin{promptbox}{GOAT-Bench}
Read the active goal in {TASK_TEXT}. Across subtasks, the 3D map and lifelong region and object records persist. Begin with fresh task-progress and spatial memory. Use the selected working set and navigation brief to decide where to search. Stay on the active floor.

Read the new goal modality before using a remembered object. A record from an earlier subtask may identify a promising room or approach. Compare that lead with the active goal and current views. If a known room is worth revisiting, select a grounded floor stance or change the view to inspect it. Use exploration candidates to enter unchecked sectors when location records and current observations provide no actionable lead.

Verify the category, the description's identifying relations, or the exact image-goal instance in current POV. For image goals, use features and neighboring layout from the reference image to find the same physical instance. Base the identification on features visible in current POV. If occlusion prevents a match, move to a safe stance with a different viewing angle. Keep a valid landmark in scene context even when visual inspection rules it out as the active target. Check proximity and an accessible target-facing stance in OVERHEAD. Approach and reobserve if a navigable gap remains. Submit the verified goal with subtask_stop, which advances immediately. Reaching the decision limit without submission records a failure and advances. Use stop after all subtasks are complete.
\end{promptbox}

\begin{promptbox}{Embodied question answering}
Answer {QUESTION} by inspecting its specified targets, room, supports, relations and requested property. Treat cropped or occluded regions as unresolved. Track checked areas, observed facts and missing evidence. Count each physical instance once. Retain views needed for comparison and replace obsolete evidence when full.

HM-EQA. A target clearly visible in OVERHEAD can support an answer even if POV misses it. If the target is too small to identify, move closer or change viewpoint. Use the visited trajectory to find an unchecked area. The controller can route to a candidate across rooms, including intermediate doorway crossings. Once a relevant target comes into view, switch to inspection. On stairs, select short OVERHEAD points on visible connected treads or landings. Save the pre-action POV/OVERHEAD pair or a labeled attached initial scan. Return the supported A-D option, using the best available evidence if an immediate answer is required.

MT-HM3D. Inspect both comparison targets against the requested reference and keep count subtotals by type. Save the pre-action view pair or labeled initial scan for later comparison. Record the specific property observed at each location so the two sightings can be compared after leaving the room. Reuse the annotation for a physical object when it appears again from another angle. Preserve useful facts before deleting the supporting image. Use the stair action at a nearby visible staircase. Return the supported A-D option, or the best-supported option when an immediate answer is required.

EXPRESS-Bench. Identify the exact target in POV and use shared landmarks in OVERHEAD to distinguish adjacent objects. Check the room, support, reference object and property named in the question. If a similar object is visible, inspect the feature that distinguishes it from the requested instance. Approach adjacent open floor when the target is localized. When no lead can be acted on, choose an unchecked room and record what still needs verification. Verify and approach the stair entrance before traversal. Use current RGB and retained visual evidence to give a concise free-form answer. Keep the target and relevant context in the final POV when feasible. At the decision limit, answer from the evidence collected.
\end{promptbox}

\begin{promptbox}{R2R-CE and RxR-CE}
Follow the {BENCHMARK} instruction in order and stop at the instructed endpoint.

Instruction: {INSTRUCTION}

Route progress. Split the instruction into ordered phases, retaining landmarks, sides, ordinal openings, stairs and the endpoint. Track the plan, completed phases, current phase and next phase from observed progress. An explicit turn can be a phase; merge stopping with the final approach. Before choosing, check current visual facts, the previous action's actual outcome and measured motion. Retain or correct progress using this evidence. A landmark leaving a partial view does not by itself invalidate completed progress. Point navigation may complete several phases; record the resulting progress rather than repeating completed turns. Distinguish approaching a doorway from crossing it. Count ordinal openings from the instructed reference using current views and route history, and interpret left/right from the arrival heading.

Local target and landmark memory. The current-target reminder is a temporary policy annotation for the current phase's next visible doorway, junction, turn point or standing place. Mark it only when current views locate the boundary, and choose actions from refreshed images rather than its saved pixel. Reaching the reminder's target alone does not complete a phase or establish the endpoint. A landmark policy annotation records a specific physical landmark explicitly named in the instruction and clearly observed now. Retain useful policy annotations across phases; skip uncertain or duplicate notes. Do not turn an inferred destination or generic doorway into an observed object landmark.

Local action. Use POV for landmark identity, OVERHEAD for nearby connected floor and BEV for longer mapped stretches. Match the same floor patch across views; historical images and overlays are context rather than navigation targets. Near entrances, turns and stopping boundaries, choose short grounded moves and inspect the resulting view before the next phase. Use supported intermediate points for a clear route, and simpler moves in uncertain passages. Check connected floor and possible unseen exits before declaring the route wrong. Retrace for a requested return, confirmed wrong turn or blockage, rather than merely failing to see a landmark. Do not rename a similar object to fit the expected route. Rotate for an instructed turn or a specific unseen passage, avoiding repeated back-and-forth confirmation. After unsupported or failed movement, reidentify floor and choose another supported approach. Use stairs only through the current stair-action contract.

Stop and response. STOP when current views and route history verify the endpoint identity and instructed position relation. Correct stale progress in the same response; do not add cosmetic centering or a needless approach. Budget exhaustion does not establish arrival.
{CURRENT_TARGET_REMINDER} {IMAGE_MANIFEST} {PUBLIC_STATE}
Return one JSON action. Begin reason with visual facts, actual outcome and route progress. On each non-STOP action, record PLAN, DONE, NOW and NEXT in task-progress memory, then supply the current action type, parameters and supported spatial memory edits. Keep PLAN tied to the instruction and distinguish NOW from completed phases.

{CURRENT_ACTION_CONTRACT}
\end{promptbox}

\end{document}